\pdfoutput=1

\documentclass[journal]{IEEEtran}

\usepackage{times}

\usepackage{soul}
\usepackage{footnote}
\usepackage{url}
\usepackage[small]{caption}
\usepackage{graphicx}
\usepackage{amsmath, bm}
\usepackage{booktabs}
\usepackage{color}
\usepackage{subfigure}
\usepackage{enumitem}
\usepackage[ruled,linesnumbered]{algorithm2e}

\definecolor{todo}{rgb}{0,0,0}
\definecolor{new}{rgb}{0,0,0}
\definecolor{duo}{rgb}{0,0,0}
\definecolor{alg}{rgb}{0,0,0}
\definecolor{mohit}{rgb}{0,0,0}

\title{Accelerating Reinforcement Learning using EEG-based Implicit Human Feedback}

\author{
\vspace*{10pt}
\IEEEauthorblockN{Duo Xu*\thanks{* co-first authors. Part of this paper has been published in AAAI Workshop on Reinforcement Learning in Games.}, \and Mohit Agarwal*, \and Ekansh Gupta, \and Faramarz Fekri, \and Raghupathy Sivakumar} \\
\IEEEauthorblockA{ 
Georgia Institute of Technology\\
 Atlanta, GA, 30332 \\
Emails: \{dxu301, mohit, egupta8, fekri, siva\}@ece.gatech.edu}
\vspace*{-10pt}
}

\begin{document}
\maketitle
\begin{abstract}
Providing Reinforcement Learning (RL) agents with human feedback can dramatically improve various aspects of learning. However, previous methods require human observer to give inputs explicitly (e.g., press buttons, voice interface), burdening the human in the loop of RL agent's learning process. Further, providing explicit human advise (feedback) continuously is not always possible or too restrictive, e.g., autonomous driving, disabled rehabilitation, etc.
In this work, we investigate capturing human's intrinsic reactions as implicit (and natural) feedback through EEG in the form of error-related potentials (ErrP), providing a natural and direct way for humans to improve the RL agent learning. As such, the human intelligence can be integrated via implicit feedback with RL algorithms to accelerate the learning of RL agent. We develop three reasonably complex 2D discrete navigational games to experimentally evaluate the overall performance of the proposed work. And the motivation of using ErrPs as feedbacks is also verified by subjective experiments. Major contributions of our work are as follows,
(i) we propose and experimentally validate the zero-shot learning of ErrPs, where the ErrPs can be learned for one game, and transferred to other unseen games, (ii) we propose a novel RL framework for integrating implicit human feedbacks via ErrPs with RL agent, improving the label efficiency and robustness to human mistakes, and (iii) compared to prior works, we scale the application of ErrPs to reasonably complex environments, and demonstrate the significance of our approach for accelerated learning through real user experiments. 

\end{abstract}
    
\section{Introduction}



  

\textcolor{new}{AI systems are increasingly applied to real-world tasks that involve interaction with humans. And humans are often in the loop of the RL agent's learning process. Self-driving cars learn with humans ready to intervene in dangerous situations. Facebook's algorithm for recommending trending news stories has humans filtering out inappropriate content}. 
Therefore RL with {\it human-in-the-loop} has inspired several research efforts where either an alternative (or supplementary) feedback is obtained from the human participant, such as human rankings or ratings \cite{el2016score}, human robot interaction and rehabilitation engineering for the disabled \cite{iturrate2010robot,knox2012learning}, or the learning is performed through human demonstrations \cite{ng1999policy}. Such approaches with explicit human input despite being highly effective, severely burdens the human interacting with RL agent. 
Further, it is difficult or even impossible to obtain the explicit human feedback in various situations, e.g., autonomous driving, disabled users, etc. 

In this work, we investigate an alternative paradigm to obtain the human feedback in an implicit manner (by tapping directly into the intrinsic brainwaves)
that substantially increases the richness of the reward functions, while not severely burdening the human-in-the-loop. We study the use of electroencephalogram (EEG) based brain waves of the human-in-the-loop to generate \textcolor{new}{the auxiliary reward functions to augment the learning of RL agent}. Such a model will benefit from the natural rich activity of a powerful sensor (the human brain), but at the same time not burden the human since the activity being relied upon is {\em intrinsic}. This paradigm is inspired by a high-level error-processing system in humans that generates error-related potential (ErrP) \cite{scheffers1996event,bentin1985event}, a negative deflection in the ongoing EEG signals.  
When a human recognizes an error made by an agent, the elicited ErrP can be captured through EEG to inform agent about the sub-optimality of the taken action in the particular state. Human feedback obtained in this manner is direct and fast while being natural and easy for humans. This widens the applicability of such RL-human interactive systems where the RL agents are deployed in the real-world environment, and increased latency of human feedback could create unwanted situations. Further, obtaining large amount of explicit feedback is infeasible due to the increased cognitive load \cite{radlinski2006evaluating}. Additionally, EEG-based feedback allows disabled users to provide the feedback, where explicit communication pathway is not available.


Previous works have \cite{chavarriaga2010learning,salazar2017correcting} demonstrated the benefit of error-potentials in a very simple setting (i.e., very small state-space, and two actions), and used ErrPs as the only reward. As a baseline contribution, we scale the feasibility of capturing error-potentials (of a human observer watching an agent learning to play games) to reasonably complex environments, and then experimentally show that decoded ErrPs can be appropriately used as an auxiliary reward function to a RL agent. In order to validate the motivation of using ErrPs as feedbacks, we make a case for using ErrPs by specifying the advantages it offers over other brain potentials and perform user studies to show that obtaining human feedback implicitly through ErrPs outperforms explicit human labeling in terms of labeling accuracy, latency of feedback and user comfort. \textcolor{new}{We also show that the \textit{full access} approach, inquiring human feedback on every state-action pair visited by RL agent, can significantly speedup the learning of the RL agent.} 

Despite the accelerated performance of \textit{full access} approach, it is not scalable to complex environments with many state-action pairs. Since the \textit{full access} relies on the implicit human feedback on each and every state-action pair, it would be
extremely inconvenient, impractical, and time consuming for the end-user (even when the user is providing the feedback without any explicit actions). Furthermore, the EEG-based implicit feedback is stochastic, (i.e., the error rate of decoding human feedback is not perfectly zero) which could possibly diverge the training of the RL agent when the feedback has relatively high error rate.

In this context, we first argue that the definition of ErrPs \textcolor{new}{can be learned in a zero-shot manner} across different environments. We experimentally validate that ErrPs of an observer can be learned to decode for a specific game, and the definition can be used as-is for another game without requiring re-learning of the ErrP decoding. This is notably different from previous approaches \cite{chavarriaga2010learning,salazar2017correcting}, where the labeled ErrPs are obtained in the same environment the RL agent is trying to solve. \textcolor{new}{We contend that previous approaches are not practical, since ErrP decoder cannot be trained and tested in the same environment.} 

We develop a framework to integrate deep RL (DRL) model with the implicit human feedback mechanism (via ErrP) in a practical, sample-efficient manner. 
Our proposed framework allows humans to provide their feedback implicitly prior to the agent training, reducing the cognitive load on humans, and hence the cost of human supervision. In the presented framework, prior to the training of RL agent, we present randomly generated demonstrations to a human for giving feedback (implicitly via ErrP), and learn an auxiliary reward function to reflect the human decision and intelligence hidden behind ErrP labels. We then pass this auxiliary reward to the RL agent to accelerate the learning process in sparse-reward environments.

Similar previous work lies in the streamline of human-agent interaction via reward shaping \cite{brown2019extrapolating,brys2015reinforcement,knox2009interactively,taylor2011integrating,warnell2018deep,xiao2020fresh}. However, the stochasticity (or erroneous nature) in the human feedback was not specifically accounted in previous methods. Hence, false ErrP labels (errors due to the collection and decoding of brainwaves) make the training less robust and unstable. 
Thus, we learn the auxiliary reward from human feedback in a way of being robust to wrong ErrP labels. It is assumed that human feedbacks coming from the optimal policy according to human intelligence. In order to tackle wrong feedbacks, we model that optimal policy as a soft-Q policy \cite{haarnoja2018soft} and learn the corresponding Q function via maximum likelihood with ErrP labels, where the probabilistic modeling has higher robustness than deterministic one. Then in order to make the learned Q function more compatible with the state space, we introduce a baseline function to smoothen that. Finally, at the RL agent side, the environmental reward (sparse) and the auxiliary reward learned from human feedback are combined to form the received reward.

We present results of real ErrP experiments to evaluate the acceleration in learning, and sample efficiency, of the proposed frameworks. \textcolor{new}{We show that such implicit feedback approach can accelerate the training of RL agent by 2.25x, while reducing the number of queries required by 75.56\%.} 
In summary, the novel contributions of our work are,
\begin{enumerate}[noitemsep, topsep=0pt, leftmargin=*]
\item We demonstrate the zero-shot learning of error-potentials over various visual-based RL problems (discrete grid-based navigation games, studied in this work), enabling the estimation of implicit human feedback in new and unseen environments without re-training of ErrP decoder. We also verify the superiority of ErrPs over manual feedback by subjective experiments, in terms
of labeling accuracy, latency of feedback and user comfort. 
\item In order to reduce the sample complexity of ErrP labels, we propose a new framework of integrating human feedback into RL via reward shaping. It is a novel approach specifically considering robustness against mistakes in human feedback. We first generate a set of random trajectories by Monte Carlo Tree Search (MCTS), balancing exploration and exploitation. Then we collect ErrP labels in experiments by demonstrating these trajectories to human observers. By learning optimal Q function with decoded labels, we derive an auxiliary reward function to augment the learning of the following RL agent.
\item We scale the implicit human feedback (via ErrP) based RL to reasonably complex environments. With subjective analysis of ErrP decoding errors and ablation study, we demonstrate the significance of our approach through experiments on various human subjects. 
\end{enumerate}

\textcolor{new}{
Our work 
demonstrates the potential of intuitive human robot interaction, facilitating robotic control by implicit human feedback in the form of ErrPs. We believe the contribution presented in this work, i.e., zero-shot learning of ErrPs and RL framework to reduce the human cognitive load, would inspire such implicit human feedback system to be deployed in practical robotic applications, such as autonomous driving or end-user applications for disabled, where explicit human feedback is not available. 
}

\section{Related Work}
The impact of feedback provided by a human to an agent in RL settings has been investigated by multiple researchers. A survey of recent research in using human guidance for deep RL tasks is presented in \cite{zhang2019leveraging}. We summarize related work in some of these techniques that are most relevant to us. In addition to rewards from environment, reward shaping learns an auxiliary reward function to accelerate the learning process of the agent \cite{chang2006reinforcement,charles2001social,thomaz2006reinforcement}. \cite{knox2009interactively} presented a framework called TAMER (Training an Agent Manually via Evaluative Reinforcement) that enabled shaping (interactively training an agent via an external signal provided by a human). Then the author extended this work to enable human feedback to augment an RL agent that learned using an MDP reward signal \cite{knox2011augmenting,knox2012learning}. Recently an architecture called Deep-TAMER \cite{warnell2018deep} has extended the TAMER framework to environments with high-dimensional state spaces. DQN-TAMER \cite{arakawa2018dqn} modeled other characteristics of human observers, such as facial expressions, from which human reward was inferred. 

Human preference \cite{christiano2017deep,wirth2016model} is another approach to communicate complex goals to allow systems to interact with real-world environments in a meaningful way. This allowed the RL agent to directly learn from expert preferences. However, this approach is limited by assumptions on the existence of a (total)
order among the set of trajectories. The author proposed a framework called Human-Agent Transfer (HAT) \cite{taylor2011integrating}. It directly used demonstrations provided by a human operator to synthesize a baseline policy, which is to guide the learning of the agent. CHAT \cite{wang2017improving} extended HAT to consider uncertainty in summarizing demonstrations and further improve the performance.

‘Potential functions’ is also used in Potential-based reward shaping (PBRS) methods to accelerate the learning process, while preserving the identity of optimal policies \cite{ng1999policy,wiewiora2003principled,xiao2019potential}. The potential function was designed to encode ‘rules’ of the environment of the RL agent. However, potential functions will typically need to be pre-specified. This has restricted the use of PBRS to tabular / low-dimensional state spaces.

In previous reward-shaping work mentioned above, human feedback is explicit, requiring active human labeling or attention, and the mistakes in human feedback are not specifically tackled. Here we propose to read implicit human feedback from error-potential hidden in human brain waves, and deal with wrong feedback in a robust approach.
Recently, there is a long line of papers studying reinforcement learning from human feedback, such as \cite{daniel2015active, el2016score, wang2016learning, wirth2016model, christiano2017deep}. However, they are only about explicit human feedback or labeling, and they all assume human feedback is noiseless. In this work, we use reward function learned by imitation learning to augment the following RL agent.

Numerous works \cite{carter1998anterior,holroyd2003errors,holroyd2002neural} have studied a high-level error-processing system in humans generating the error-related potential/negativity (ErrP or ERN). 

Interaction, response, and feedback ErrPs have been heavily investigated in the domain of choice reaction tasks, where human is actively interacting with the system \cite{schalk2000eeg,blankertz2003boosting,parra2003response,ferrez2005you,ferrez2008error} and the error is made either by the human or by the machine. \cite{kim2017intrinsic} demonstrated the use of ErrP signals in an interactive RL task, when the human is actively interacting with the machine system. \cite{ferrez2005you} explored the ErrPs when human is silently observing the machine actions (and does not actively interact). Works at the intersection of ErrP and RL \cite{chavarriaga2010learning,salazar2017correcting} demonstrate the benefit of ErrPs in a very simple setting (i.e., very small state-space), and use ErrP-based feedback as the only reward. Moreover, in all of these works, the ErrP decoder is trained on a similar game (or robotic task), essentially using the knowledge that is supposed to be unknown in the RL task. In our work, we use labeled ErrPs examples of very simple and known environments to train the ErrP decoder, and integrate ErrP with DRL in a sample-efficient manner for reasonably complex environments. 

\section{Preliminaries and Setup}
\label{sec:prelim}
\subsection{RL and the Q-function}
\textbf{Definitions:}
Consider a Markov Decision Process (MDP) problem $M$, as a tuple $<\mathcal{X}, \mathcal{A}, P, P_0, R, \gamma>$, with state-space $\mathcal{X}$, action-space $\mathcal{A}$, transition kernel $P$, initial state distribution $P_0$, accompanied with reward function $R$, and discounting factor $0\le\gamma\le 1$. In this work, we only consider MDP with discrete actions and states. In model-free RL method, the central idea of most prominent approaches is to learn the Q-function by minimizing the Bellman residual, i.e., $\mathcal{L}(Q)=\mathcal{E}_{\pi}\big[\big(Q(x,a)-r-\gamma Q(x',\hat{a})\big)^2\big]$, and temporal difference (TD) update where the transition tuple $(x,a,r,x')$ consists of state, action, reward and next state respectively, i.e., a consecutive experience under behavior policy $\pi$. 

\noindent 
\textbf{Bayesian Deep Q Network:}
\label{subsec:bdqn}
The Q function model adopted in this paper is {\it Bayesian DQN} \cite{azizzadenesheli2018efficient}. It is a neural architecture where the Q-function is approximated as a linear function, weighted by $\omega_a, a\in\mathcal{A}$, of the feature representation of states $\phi_{\theta}(x)\in\mathcal{R}^d$, parameterized by neural network with weights $\theta$. The weights $\omega_a$ follow the Gaussian distribution from Bayesian linear regression.

\subsection{Games and EEG}
Using a game as a proxy for a real-life environment is  beneficial  in  the  context  of  human  assisted RL algorithms. Games are a fertile ground for the definition, understanding, and improvement of RL algorithms in a low overhead and speedy fashion. Games have now evolved to help understand the world around us and make optimal strategies to tackle various difficult and high-risk real-world situations. For example, Foldit is an online puzzle video game about protein folding. The users of the game helped to solve the structure of a protein-sniping enzyme critical for reproduction of the AIDS virus. A curious planet with four stars was discovered through another game, Planet Hunter, along with the discovery of 40 other planets with the potential of having life-forms \cite{gatech_lab}. Motivated by these  studies,  we  use games  as environments for gathering ErrP data from humans in order to accelerate our RL algorithm. 

Specifically, we use electroencephalogram (EEG) signals to generate the implicit feedback that can be used by the RL algorithm. Electroencephalography (EEG) is a mechanism to  detect electrical activity in a human brain using small, metal discs (electrodes) attached to the scalp. Brain cells communicate via electrical impulses and EEG helps in recording this activity. EEG is growing to be a bonafide and easy to use \cite{chi2020paper} input modality in several applications such as communication \cite{Agarwal2015THINKTP} \cite{blink_gupta}, lifestyle \cite{cerebro_p1},
RL \cite{wearsys_hitl} etc. and due to the wider availability of EEG headsets off-the-shelf,
access to a user’s EEG data is easier than it has ever been. We use EEG because error potentials, which are an outcome of a high level error-performance system,  manifest themselves as a negative deflection in the EEG signal activity of a human \cite{scheffers1996event,bentin1985event,nieuwenhuis2001error}. This is then detected and used into the reward function of the reinforcement learning algorithm. We also make the case for using error potentials over manual labeling in section \ref{sec:errp_case}.

\subsection{System Setup and Data Collection}
\label{subsec:experiment}
\subsubsection{BCI for implicit feedback}
We designed and developed an experimental protocol, where a machine agent plays a computer game, while a human silently observes (and assesses) the actions taken by the machine agent. These implicit human reactions are captured by placing raw electrodes on the scalp of the human brain in the form of EEG potentials. The electrode cap (BIOPAC CAP-100C) was attached with the OpenBCI Cyton\footnote{http://openbci.com} platform, which was further connected to a desktop machine over the wireless channel.
In the game design (developed on OpenAI Gym), we open a TCP port, and continuously transmit the current state-action pair using the TCP/IP protocol. We used OpenViBE software \cite{renard2010openvibe} to record the human EEG data. OpenViBE continuously listens to the TCP port (for state-action pairs), and timestamps the EEG data in a synchronized manner. We recruited a total of five human subjects (mean age 26.8 with standard deviation of 1.92, 1 female)
using standard procedures with their consent. For each subject-game pair, the experimental duration was less than 15 minutes. The agent took action every 1.5 seconds during the experiment. The University Institutional Review Board reviewed and approved all the research protocols for the user data collection. 

\subsubsection{The Games}
\label{subsec:gameenv}
\begin{figure}
\centering
\subfigure[Game Environments]{
\begin{minipage}[t]{0.65\columnwidth}
\centering
\includegraphics[width=0.75\columnwidth]{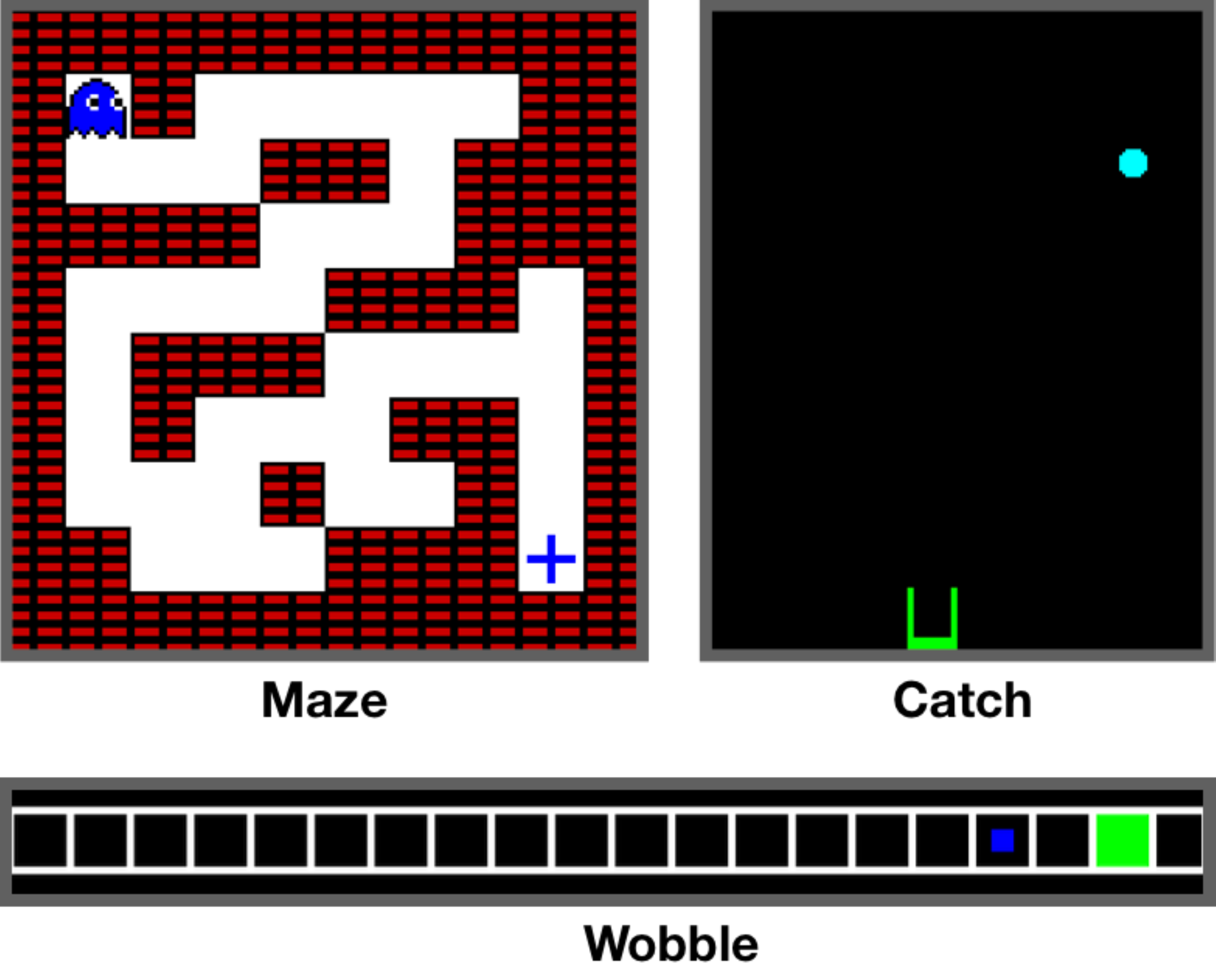}
\label{fig:games_screenshot}
\end{minipage}
}

\subfigure[Experiment Bench]{
\begin{minipage}[t]{0.65\columnwidth}
\centering
\includegraphics[width=0.75\columnwidth]{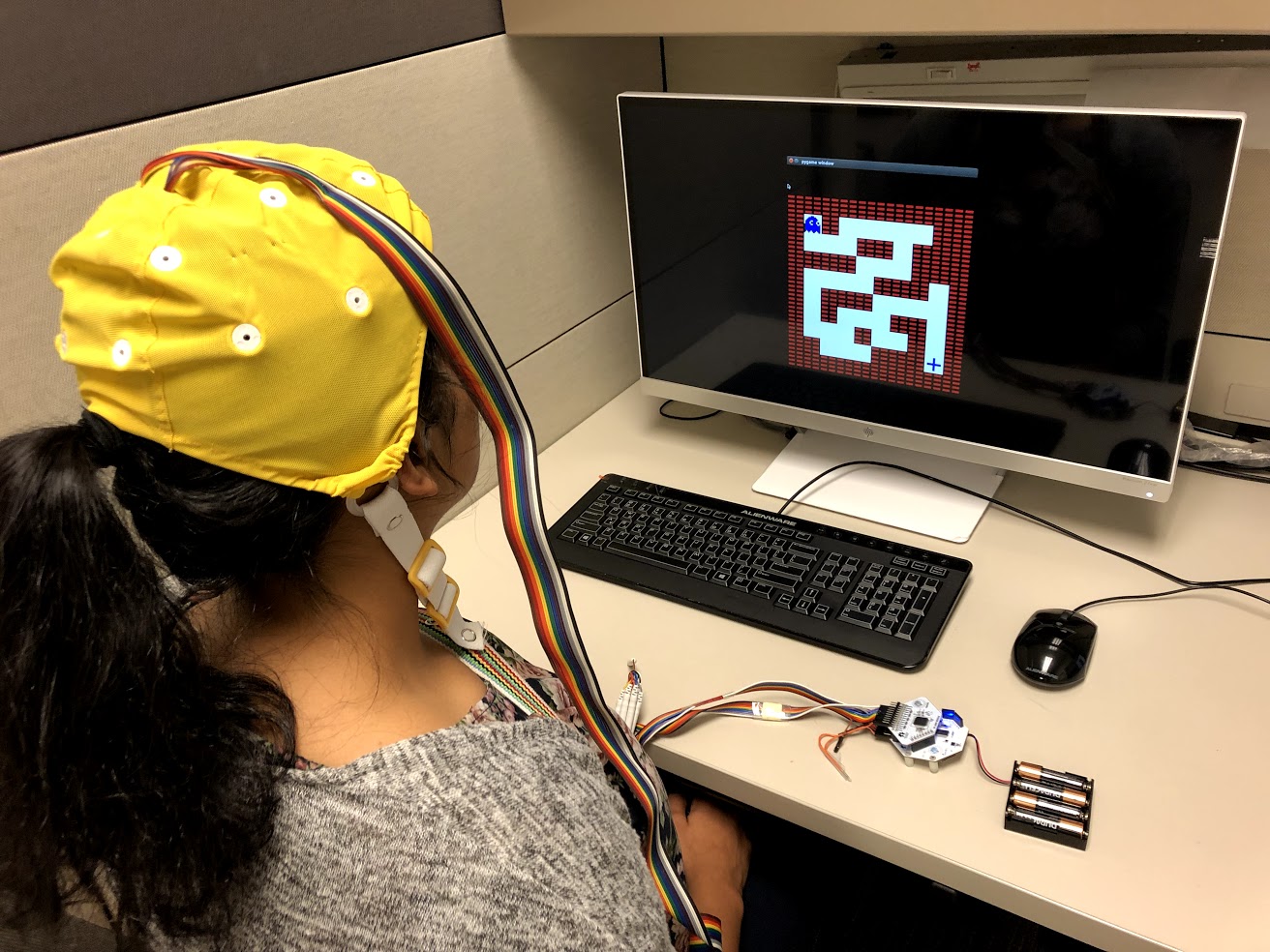}
\label{fig:expbench}
\end{minipage}
}
\vspace*{-10pt}
\caption{Experimental framework}
\label{fig:gamesnbench}
\end{figure}

We have developed three discrete grid-based navigational games in OpenAI Gym \textit{Atari} framework, namely Wobble, Catch, and Maze (Fig. \ref{fig:games_screenshot}). 
\newline
\textbf{Wobble:} Wobble is a simple 1-D cursor-target game, where the middle horizontal plane is divided into 20 discrete blocks. At the beginning of the game, the cursor appears at the center of the screen, and the target appears no more than three blocks away from the cursor position. The action space constitutes moving to the left or right. 
The game is finished when the cursor reaches the target. Once the game is finished, a new game is started with the cursor in place. \newline 
\textbf{Catch:} Catch is a simplistic version of \textit{Eggomania}\footnote{https://en.wikipedia.org/wiki/Eggomania} (Atari 2600 benchmark), where we display a single egg on the screen at a time. The has a 10x10 grid, where the \textit{egg} and the \textit{cart}, both occupy one block. The action space of the agent consists of \textit{NOOP} (no operation), \textit{left} and \textit{right}. At the start of the game, the horizontal position of the egg is chosen randomly. At each time step, the \textit{egg} falls one block in the vertical direction. \newline 
\textbf{Maze:} Maze is a 2-D navigational game, where the agent has to reach a fixed target (shown with a \textit{plus} symbol). The screen is divided into 10x10 square blocks. The action space consists of four directional movements. The only reward here is the result of the episode, i.e., win or lose. If an agent moves, but hits a wall, a quick blinking of the agent is displayed, to render the action taken by the agent.


\color{mohit}




\section{The Case for Error Potentials}
\label{sec:errp_case}

We make the case for using error potentials (ErrP) by first, touching upon some of their characteristic advantages that have been documented in academic literature, and then by providing experimental evidence of the superiority of ErrPs over manual or explicit labeling. We show that using ErrPs as feedbacks reduces the cognitive burden of the subjects, and provides better trade-off between latency and correctness of the feedback. \\
\subsection{The usefulness of ErrPs for error detection}
To begin with, relying on error-potentials provides two primary benefits:
\begin{enumerate}[label=(\alph*)]
    \item \textbf{Provides a generalized notion of error-detection:} Error-potentials are elicited when a user is presented with an incongruent (or erroneous) stimulus in a diverse set of tasks \cite{gehring_1995} implying that the error-processing system is generic (i.e., not specific to the task or sensory organ). Error-potentials are observed across a wide variety of input modality (e.g., audio \cite{Falkenstein1991EffectsOC}, visual \cite{errp_2000}, somatosensory \cite{errp_1997}, etc.). {This is in contrast to other elicited potentials in the brain which cater to the stimuli of a specific category.} For instance, the P600, N300, P300, and N200 are elicited when a subject is presented with syntactic anomalies in sentences \cite{Osterhout2002}, semantically inconsistent word and picture pairs \cite{MAGUIRE20131}, interruption of a stimulus with another divergent stimulus \cite{p300_2}, and detection of mismatch in a stimulus \cite{n200_1} respectively. 
    {Thus, the generalized mechanism for eliciting ErrPs is one of the characteristic advantage that it offers, unlike other brain-potentials  specific to a stimulus or modality.}
    \item \textbf{Evolutionary Significance:}
    \label{subsec:errp_evo}
    Error-potentials in primates are {well-founded} and universal (exhibiting similar behaviours across individuals) as they have an evolutionary significance due to their importance in cognition, learning, and survival. Error-potentials enable the learning process via the administration of rewards and punishments in Anterior Cingulate Cortex (ACC) \cite{Holroyd2002TheNB}. In monkeys, error-potentials were generated in anterior cingulate sulcus, when monkeys made errors in a simple response task. \cite{niki_1979} found error-recognition units in monkeys’ anterior cingulate sulcus that were activated when the animals received negative feedback in the form of absence of an expected reward. Similarly, \cite{Gemba_1986} found that when monkeys made errors in a simple response task, error-related potentials were generated in the anterior cingulate sulcus, thereby advocating that ErrPs link human and non human primates on the basis of error monitoring. {Universality of ErrPs guarantees that it occurs naturally in humans and the evolutionary importance of ErrPs in learning points toward them being a foundational element in the human cognition, enabling their use without worrying about individual differences based on learned behavior.}
    
\end{enumerate}

\subsection{Empirical motivation for using intrinsic error-potentials over manual labeling}
\label{subsec:Motivation}
\textbf{Experimental Methodology:} We conducted an experiment in which we asked subjects to label the actions of an AI agent in a maze. If the agent took a correct action, they needed to press a certain key and if the agent performed a wrong action, they were supposed to press another key. We conducted this experiment to find differences between manual labeling and labeling using EEG experiments in terms of user comfort and labeling accuracy. The methodology of this experiment was simple. We designed a maze game and generated 3 instances of it where each instance got progressively faster (to study the impact of time pressure on mental comfort and accuracy). The first instance had a time delay of 1.5 seconds between successive actions of the agent while the second and the third had a delay of 1.0 and 0.5 seconds respectively (We use these delay values as they lie around the latency value we have used in the EEG experiments and they also help us know the variation of manual labeling accuracy with respect to latency). We made the subjects play 3 trials in each instance (thus totaling to 9 trials overall) where we randomized the order of instances to avoid biasing the users to a particular order of the game. In all these instances, the AI agent made the correct move with a probability of 0.8. Once subjects finished playing 3 trials of an instance, we redirected them to a Qualtrics survey where they had to provide their subjective feedback about the experiment. Thus, there were 3 forms that each subject had to fill (one per instance). We used Amazon's Mechanical Turk to request anonymous workers to complete this task. The Institutional Review Board approved the study protocol.

\textbf{Results:} We obtained a total of 281 responses (87, 91, and 103 unique user responses for the 1.5s, 1.0s, and 0.5s instances of the game respectively). On average, for the 1.5s instance of the game, we obtained a true positive rate of 56.6\% and 41.5\% for correct and incorrect actions of the maze agent respectively. We also obtained a feedback latency of 376ms and 540ms for correct and incorrect actions of the maze agent respectively. Note that correct and incorrect actions of the agent corresponds to the non-Errp and ErrP respectively, during EEG experiments. For the 1.0s instance of the game, we obtained a true positive rate of 49.8\% and 38.8\% for correct and incorrect actions of the maze agent respectively. We also obtained a feedback latency of 288ms and 456ms for correct and incorrect actions of the maze agent respectively. For the 0.5s instance of the game, we obtained a true positive rate of 34.9\% and 14.1\% for correct and incorrect actions of the maze agent respectively. We also obtained a feedback latency of 179ms and 207ms for correct and incorrect actions of the maze agent respectively. However some trials in these experiments had very poor labeling rate (some subjects only labeled less than 50\% of the actions available). In order to prevent the results from being swayed by inert participants, we decided to separate them from the active participants.

\begin{figure}
\centering
\begin{minipage}[t]{\columnwidth}
\centering
\includegraphics[width=0.99\columnwidth]{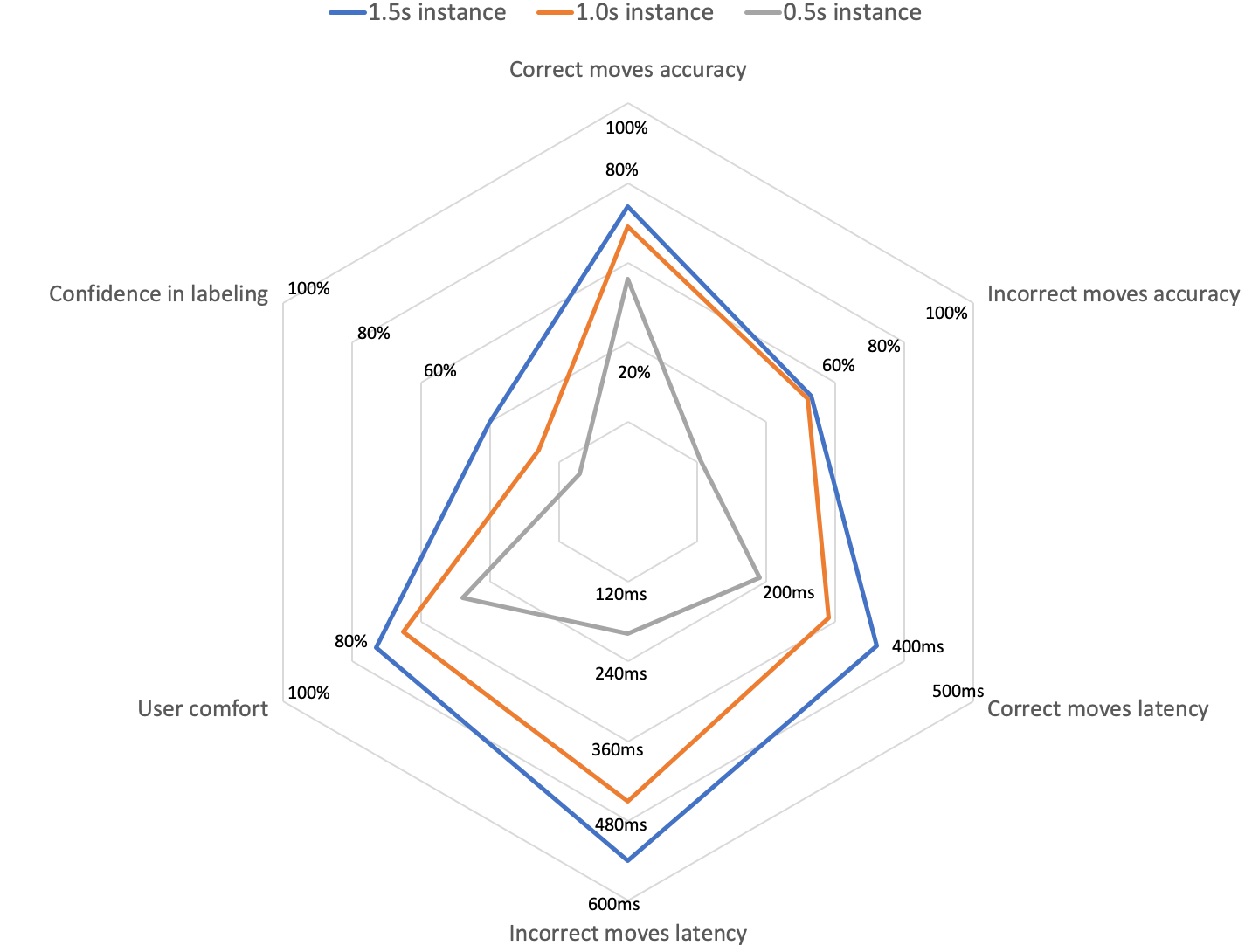}
\end{minipage}
\caption{Difference between instances of the maze game used for manual labeling}
\label{fig:maze_radar}
\end{figure}

\begin{figure}
\centering
\begin{minipage}[t]{\columnwidth}
\centering
\includegraphics[width=0.8\columnwidth]{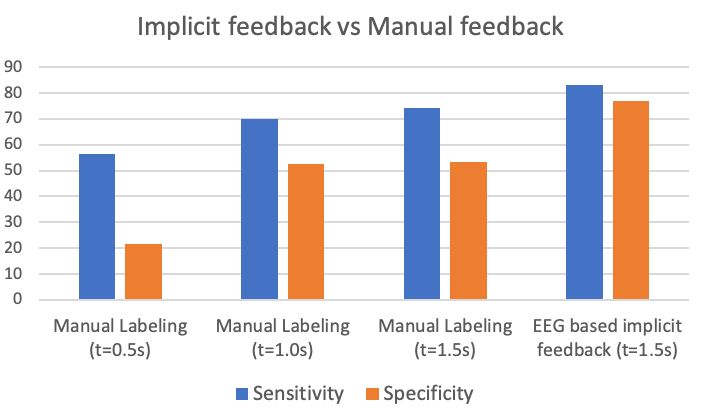}
\end{minipage}
\caption{Manual labeling compared with implicit feedback using EEG on maze game}
\label{fig:eeg_manual}
\end{figure}

We decided to remove the trials for the users which had less than 50\% feedback rate. In other words, we removed the trials where participants failed to provide the feedback for at least 50\% of all the actions. This led to the removal of 22 users from the 1.5s instance of the game (25\%), 28 users from the 1.0s instance of the game (31\%), and 44 users from the 0.5s instance of the game (43\%). After this filtering, for the 1.5s instance of the game, we obtained a true positive rate of 74.1\% and 53.4\% for correct and incorrect actions of the maze agent respectively. We also obtained a feedback latency of 364ms and 539ms for correct and incorrect actions of the maze agent respectively. For the 1.0s instance of the game, we obtained a true positive rate of 69.8\% and 52.6\% for correct and incorrect actions of the maze agent respectively. We also obtained a feedback latency of 290ms and 451ms for correct and incorrect actions of the maze agent respectively. For the 0.5s instance of the game, we obtained a true positive rate of 56.4\% and 21.6\% for correct and incorrect actions of the maze agent respectively. We also obtained a feedback latency of 177ms and 203ms for correct and incorrect actions of the maze agent respectively. These values are summarized in Table \ref{tab:maze_manual_label} and are contrasted pictorially in Fig \ref{fig:maze_radar}. The results of manual labeling are also contrasted with those of implicit feedback using EEG in Fig \ref{fig:eeg_manual}.

\begin{table}
\vspace{-5pt}
\caption{Accuracy and latency for maze game manual labeling}
\setlength{\tabcolsep}{1.2mm}{\footnotesize
\begin{tabular}{|c|c|c|c|c|c|}
\toprule
Time&Subjects & Correct  & Incorrect  & Correct  & Incorrect  \\
Interval (s)& & Moves  & Moves  & Moves  & Moves  \\

&  & TPR (\%)  &TPR (\%) & latency (ms) & latency (ms) \\
\midrule
1.5&87 &74.1 &53.4 &364 &539 \\

1.0&91 &69.8 &52.6 &290 &451 \\

0.5&103 &56.4 &21.6 &177 &203 \\
\bottomrule
\end{tabular}}
\label{tab:maze_manual_label}
\end{table}

\textbf{Insights:}
As we can clearly see, the accuracy values for correct and incorrect actions both decrease with decrease in time interval. The labeling accuracy for correct actions seems to be more than that of incorrect actions. Both, the accuracy for correct as well as incorrect actions decreases as the time latency is decreased (thereby increasing time pressure). Even the best possible accuracy for incorrect actions is about 53.4\% (only marginally better than random labeling). This performs rather poorly compared to the labeling accuracy using ErrP.

Based on the qualitative survey responses, on a scale of 1 to 7, the users gave the 1.5s instance of the game an average comfort rating of 5.4 which declined to 4.9 and 3.9 for the 1.0s instance and 0.5s instance respectively. On being asked if they were able to mark all actions correctly, 40\% of the subjects answered in the affirmative in the 1.5s instance of the game, which declined to 26\% and 14\% in the 1.0s and the 0.5s instance of the game. Across the board, the majority of the participants reported that the ideal time interval for them to correctly label all actions of the agent would be between 1.5s and 3.0s or larger. 64\% of the participants in the 1.5s instance of the game reported that reducing the time interval of the game to 1.0s would decrease their labeling accuracy, and 69\% of the participants reported that it would increase their mental burden. 52\% of the participants in the 1.0s instance of the game reported that reducing the time interval of the game to 0.5s would decrease their labeling accuracy, and 60\% of the participants reported that it would increase their mental burden. In contrast, 64\% of the participants in the 1.0s instance of the game reported that increasing the time interval from 1.0s to 1.5s would increase their labeling accuracy and decrease their mental burden. 49\% of the participants in the 0.5s instance of the game reported that reducing the time interval of the game further would decrease their labeling accuracy, and 53\% of the participants reported that it would increase their mental burden. To summarize, the users felt increasing discomfort and cognitive burden as the time latency reduced from 1.5s to 1.0s and further to 0.5s. They also reported that the optimal time latency for comfortable manual labeling would be between 1.5s and 3s. This was also evident from the fact that more than 60\% of the participants anticipated reduction in their accuracy if time latency was to be decreased from 1.5s.

\color{black}
\section{Integrating RL with Implicit Human Feedback: A Naive Approach}
\label{sec:full_access}

In this section, we provide our baseline contribution, i.e., (i) we demonstrate the feasibility of capturing error-potentials of a human subject watching an RL agent learning to play several different games, and then decoding the human feedback (judgment) on the observed state-action pair appropriately, and (ii) using them as an auxiliary reward function to accelerate the learning of the RL agent.
\subsection{Obtaining the Implicit Human Feedback: Decoding ErrPs}
\label{sec:errp_algo}

In order to obtain the implicit human feedback, we need to detect the presence or absence of ErrPs inside the EEG waveform. This requires training a model that can interpret the EEG signal of a human and classify it as ErrP or non-ErrP correctly. EEG signals are inherently very noisy, and when combined with external factors like improper electrode placements, variance across users pose significant challenges in the reliable estimation of error-potentials.

We rely on the Riemannian Geometry framework for the classification of a human's intrinsic reaction \cite{barachant2014plug}.
 This framework is state-of-the-art for detecting any event-related potentials, and provides two primary advantages over other classifiers:
 \begin{itemize}
     \item The estimation algorithm operates in signal space (rather than source space), and hence minimizes the distortions due to the electrode placements.
     \item The spatial filtering algorithm maximizes the signal to signal plus noise ratio (SSNR) to mitigate the interference and noise.
 \end{itemize}
 
 We consider the classification of error-related potentials as a binary classification task indicating the presence (i.e., action taken by the agent is incorrect) and absence of error (i.e., action taken by the agent is correct). 
 We bandpass filtered the raw EEG data within [0.5, 40] Hz. 
We then extracted Epochs of 800ms relative to pre-stimulus 200ms baseline and subjected them to spatial filtering. In spatial filtering, we compute prototype responses of each class, i.e., ``correct'' and ``erroneous'', by averaging all training trials in the corresponding classes(``xDAWN Spatial Filter'' \cite{rivet2009xdawn,congedo2013new}). ``xDAWN filtering'' projects the EEG signals from sensor space (i.e., electrode space) to the source space (i.e., a low-dimensional space constituted by the actual neuronal ensembles in brain firing coherently).
We then compute the covariance matrix of each epoch, and concatenate it with the prototype responses of the class. Further, we select relevant channels through backward elimination \cite{barachant2011channel} to achieve dimensionality reduction and project
the filtered signals on the tangent space \cite{barachant2013classification} for feature extraction. We then normalize (using L1 norm) the obtained feature vector and feed it to a regularized regression model. We select a threshold value for the final decision by maximizing accuracy offline on the training set. We present the algorithm to decode the ErrP signals in Algorithm \ref{algo:errp}.
\begin{algorithm}
    \label{algo:errp}
    \caption{Riemannian Geometry based ErrP classification algorithm}
    \label{algo:errp}
    \SetKwInOut{Input}{Input}
    \SetKwInOut{Output}{Output}

    \Input{raw EEG signals $EEG$}
    Pre-process raw EEG signals \;
    Spatial Filtering: xDAWN Spatial Filter ($nfilter$) \;
     Electrode Selection: ElectrodeSelect ($nelec$, metric='riemann') \;
    Tangent Space Projection : TangentSpace(metric = ``logeuclid'')
    Normalize using L1 norm \;
    Regression: ElasticNet \;
    Select decision threshold by maximizing accuracy
\end{algorithm}


\subsection{The Full Access Method}
\label{subsec:full_access}
A naive approach \textcolor{new}{to integrate the human feedback with RL models is reward shaping with \textit{full access}. We obtain human feedback on every visited state-action pair while RL agent is learning, and add a negative penalty to the environmental reward in case an ErrP is detected.  
We present the evaluation result of this method based on real ErrP data later in the evaluation section (section \ref{sec:resultsNaive}), validating that \textit{full access} method can significantly accelerate the learning of the RL agent.
However, obtaining the human feedback for every state-action pair is time-intensive and not practically feasible.} \textcolor{new}{In the next section, we provide our novel contributions to practically obtain and integrate the implicit feedback with the learning of RL agent.} 



Interaction, response, and feedback ErrPs have been heavily investigated in the domain of choice reaction tasks, where human is actively interacting with the system \cite{schalk2000eeg,blankertz2003boosting,parra2003response,ferrez2005you,ferrez2008error} and the error is made either by the human or by the machine. \cite{kim2017intrinsic} demonstrated the use of ErrP signals in an interactive RL task, when the human is actively interacting with the machine system. \cite{ferrez2005you} explored the ErrPs when human is silently observing the machine actions (and does not actively interact). Works at the intersection of ErrP and RL \cite{chavarriaga2010learning,salazar2017correcting} demonstrate the benefit of ErrPs in a very simple setting (i.e., very small state-space), and use ErrP-based feedback as the only reward. Moreover, in all of these works, the ErrP decoder is trained on a similar game (or robotic task), essentially using the knowledge that is supposed to be unknown in the RL task. In our work, we use labeled ErrPs examples of very simple and known environments to train the ErrP decoder, and integrate ErrP with DRL in a sample-efficient manner for reasonably complex environments. 

\section{Toward Smarter Integration of RL with Implicit Human Feedback}
\label{sec:sec4}

In this section, we propose two approaches 
to \textcolor{new}{enable the deployment of ErrP-augmented RL into practical systems}. Firstly, we show that we can learn the ErrPs of an observer in a zero-shot manner, i.e. We can train an ErrP decoder for a specific game, and use the trained decoder as-is for another game without re-training the ErrP decoder. \textcolor{new}{To combat with the practical issues with obtaining ErrP labels for every state-action pairs, we propose an RL framework (motivated by imitation learning approaches) allowing humans to provide their feedback on a few trajectories prior to the learning of the RL agent. This dramatically reduces the number of feedback labels required from the human observer. }

\subsection{Robust Reward Shaping using Human Feedback}
\label{sec:second}
RL algorithms deployed in the environment with sparse rewards demand heavy explorations (require a large number of trial-and-errors) during the initial stages of training. \textcolor{duo}{In such environments, using human feedback can be very efficient for accelerating the learning process.} Previous work on reward shaping with human feedback \cite{brys2015reinforcement,knox2009interactively,taylor2011integrating,warnell2018deep,xiao2020fresh} build a specific model to generalize human feedback in state space, without tackling wrong feedback. Inspired by soft Q policy \cite{haarnoja2018soft}, we develop a novel framework of learning the auxiliary reward from human feedback to accelerate the training of the RL agent, with robustness to mistakes in ErrP labeling. 

\begin{figure}
\centering
\includegraphics[width=0.98\columnwidth, height=0.28\columnwidth]{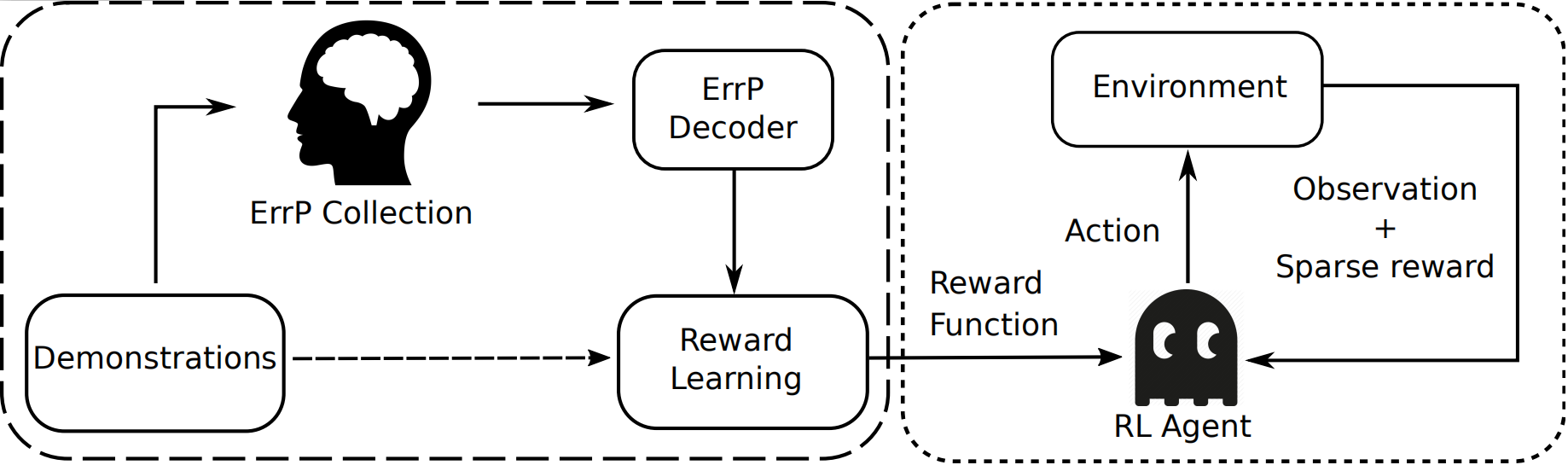}
\caption{Robust Reward Shaping with Human Feedback. The dashed arrow shows trajectories in $\mathcal{D}\cup\mathcal{D}_R$ are all used in reward learning}
\label{fig:imperfect}
\vspace*{-10pt}
\end{figure}

\textcolor{new}{In this framework, we require implicit human feedback via ErrP on all state-action pairs along trajectories (demonstrations) randomly generated initially. Before RL agent starts learning, we ask the human subjects to observe a number of trajectories, and record their implicit feedbacks in the form of ErrP on corresponding state-action pair in a dataset. Then we learn the auxiliary reward function $r_a(\cdot, \cdot)$ from these trajectories labeled by human feedback. During the RL training, the learned reward function acts as a proxy for the human feedback, compensating the sparse reward from the environment. The flowchart of the proposed learning framework is shown in Figure \ref{fig:imperfect}.} \textcolor{new}{Different from the naive baseline {\it full access} method discussed earlier, in this approach,} we require the queries for human feedback (ErrP labeling) only on trajectories generated initially, instead of querying every learning step during the training. Hence, the \textcolor{duo}{total number of ErrP queries are reduced significantly,  further reducing the load for the human in the loop}.


\noindent 
\textbf{Trajectory Generation:}
Constraint by the coherence requirement in EEG experiments, the trajectories for ErrP labeling have to be complete, containing every state-action pair from the beginning to the end of the game. Further, the selected trajectories have to cover state space as much as possible, and cannot be too far away from the optimal solutions. This is essentially the trade-off between exploitation and exploration. So we propose to use Monte Carlo Tree Search (MCTS) \cite{browne2012survey,kocsis2006bandit,silver2010monte} to generate random trajectories for ErrP experiments. It is to tackle exploration-exploitation trade-off by Upper Confidence Bound (UCB) method \cite{kocsis2006bandit}, and does not require to know the optimal solution a priori. MCTS is a general game playing technique with recent success in discrete, turn-based, and non-deterministic game domains. We choose MCTS as a trajectory sampling algorithm for its proven high-level performance, domain generality, and variable computational bounds. 

In MCTS, there is one node in the tree for each state $s$, containing a value $Q(s,a)$ and a visit count $N(s,a)$ for each action $a$, and an overall count $N(s)=\sum_aN(s,a)$. Each node is initialised to $Q(s,a)=0, N(s,a)=0$. The value is estimated by the mean return from $s$ in all simulations where action $a$ was selected in state $s$, and the only reward $r$ here is the result of the game, i.e., 1 for winning and 0 for losing. At each state $s$ of the trajectory, the action is selected to be the maximizer of the objective $Q(s,a)+c\sqrt{\frac{\log N(s)}{N(s,a)}}$, where $c$ is to trade off between reaching the target and exploring more state space \cite{kocsis2006bandit}. By the end of generating each trajectory, the return is back-propagated into Q values along the trajectory, i.e., $Q(s_t, a_t) := r + \gamma Q(s_{t+1}, a_{t+1})$. Only first $K$ generated trajectories are used in ErrP experiments. 

In experiments for collecting ErrPs, the human subject provides implicit feedback (via ErrP) over all the generated trajectories, labeling every state-action pair as a {\it positive} (if a correct action according to perceived human intelligence) or a {\it negative} sample. With decoded ErrP labels over trajectories as input, we propose a novel reward shaping method to incorporate ErrP labels into the reinforcement learning framework. It specifically tackles the problem of robustness against wrong ErrP labels, with details explained in the following section.

\noindent
\textbf{Reward Learning}
\label{sec:reward}
Since implicit human feedback via ErrP is noisy, different from previous work \cite{brys2015reinforcement,knox2009interactively,taylor2011integrating,warnell2018deep,xiao2020fresh}, instead of modeling human feedbacks by neural networks directly, we assume that human give feedbacks according to his probabilistic policy, modeled by a soft-Q policy with Q function $Q_{h,\theta}$ (with weights $\theta$), under the max-entropy principle \cite{ziebart2010modeling}. Our method is to define operators first going from demonstrations labeled by human feedbacks to the optimal Q function $Q_{h,\theta}$, and then from that Q function to the auxiliary reward function $r_a(\cdot)$. Here we assume that optimal Q function defines a soft-Q policy \cite{haarnoja2017reinforcement,haarnoja2018soft}, and learn it by solving a probabilistic classification problem via maximizing log-likelihood of human feedbacks. Following the principle of maximum entropy \cite{ziebart2010modeling}, the soft-Q policy giving human feedbacks and the corresponding value function can be expressed as follows, 
\begin{eqnarray}
\pi_{h,\theta}(\bm{a}|\bm{s})&=&\exp((Q_{h,\theta}(\bm{s},\bm{a})-V_{h,\theta}(\bm{s}))/\alpha), \nonumber \\ 
V_{h,\theta}(\bm{s})&=&\alpha\log\sum_a\exp(Q_{h,\theta}(\bm{s},\bm{a})/\alpha) \label{maxent}
\end{eqnarray}
where $\alpha$ is a free parameter, tuned empirically. Define positive samples as state-action pairs with correct labels in human feedbacks while negative samples as those with wrong labels. According to the maximum entropy principle \cite{ziebart2010modeling}, the likelihood of positive and negative samples are denoted as $\pi_{h,\theta}(\bm{a}|\bm{s})$ and $1-\pi_{h,\theta}(\bm{a}|\bm{s})$, respectively. When trajectories and human feedbacks (ErrP labels) are ready, we learn $Q_{h,\theta}(\cdot,\cdot)$ by maximizing the likelihood of both positive and negative state-action pairs in the trajectories, which is to maximize the objective \eqref{j1} in Algorithm 2, where the binary variable $\text{ErrP}(s,a)$ denotes the human feedback label. To derive auxiliary reward from the learned Q function, a naive choice is the Bellman difference, i.e., $Q_{h,\theta}(\bm{s},\bm{a})-\gamma\max_{\bm{a}}Q_{h,\theta}(\bm{s},\bm{a})$. However, due to the scarce of ErrP labels on exact state-action pairs, the function $Q_{h,\theta}$ learned by maximum likelihood may not have the shape compatible with the state dynamics of the target MDP (environments in experiments). And the derived auxiliary reward function can destabilize the learning process of RL agent.

In order to refine the reward shape and attenuate the gradient variance, we introduce another baseline function $t_{\phi}(\bm{s})$ only dependant on the state, to incorporate the state transition information, parametrized by $\phi$. Hence, the Q function is augmented as $Q_B(\bm{s},\bm{a}):=Q_{h,\theta}(\bm{s},\bm{a})+t(\bm{s})$. It can be proved that $Q_B(\cdot,\cdot)$ and $Q_{h,\theta}(\cdot,\cdot)$ induce the same optimal policy \cite{ng1999policy}. The baseline function $t^*(\cdot)$ can be learned by optimizing $t^*=\arg\min_{\phi}J_2(\phi)$, defined in \eqref{j2}, where the loss function $l(\cdot)$ is chosen to be $l_1$-norm via empirical evaluations .

For learning function $t_{\phi}(\cdot)$, in addition to the demonstration $\mathcal{D}$ in Figure \ref{fig:imperfect}, we incorporate another set of demonstrations $\mathcal{D}_R$, containing transitions that are randomly sampled from environment without the reward information. The set $\mathcal{D}_R$ is to help the learned auxiliary reward function to incorporate the state dynamics information, without the need of any human labeling. After learning both $Q_{h,\theta}(\cdot,\cdot)$ and $t_{\phi}(\cdot)$, for any transition tuple $(\bm{s}, \bm{a}, \bm{s}')$, the auxiliary reward function can be represented as
\begin{equation}
r_a(\bm{s}, \bm{a})=Q_{h,\theta}(\bm{s}, \bm{a})+t_{\phi}(\bm{s})-\gamma\max_{\bm{a}'\in\mathcal{A}}[Q_{h,\theta}(\bm{s}',\bm{a}')+t_{\phi}(\bm{s}')] \label{aux_rew}
\end{equation}
This $r_a$ is then used to augment the RL agent. In order to further attenuate the negative influence of wrong ErrP labels, when combining environmental reward $r_e$ and auxiliary reward $r_a$, we propose a coefficient $\beta(e)$, exponentially decreasing in terms of training episodes $e$, i.e., $\beta(e):=a e^{-e/b}$. Finally, the RL agent receives the shaped reward in the form of $r_e(\bm{s}_t,\bm{a}_t)+\beta(e)r_a(\bm{s}_t,\bm{a}_t)$. Empirically, the best coefficient function is $\beta(e)=3e^{-e/80}$ in experiments. 



\begin{algorithm}
    \caption{Robust Reward Shaping with Human ErrP}
    \SetKwInOut{Input}{Input}
    \SetKwInOut{Output}{Output}

    \Input{Trajectories Given Initially}
    Conduct EEG Experiments for human ErrP to label the state-action pairs along trajectories\;
    With ErrP data collected, use Algorithm \ref{algo:errp} to decode ErrP labels, i.e., $\text{ErrP}(\cdot,\cdot)$\;
    Initialize the Q function $Q_{h,\theta}(\cdot,\cdot)$ and baseline $t_{\phi}(\cdot)$\;
    Learn $Q_{h,\theta}(\cdot,\cdot)$ by optimizing $\theta$
    \begin{eqnarray}
        J_1(\theta)&:=&\sum_{(s,a)\in\mathcal{D}}\pi_{h,\theta}(a|s)(1-\text{ErrP}(s,a)) \nonumber \\
        &&+(1-\pi_{h,\theta}(a|s))\text{ErrP}(s,a) \label{j1}
    \end{eqnarray}
    where the relationship between $\pi_{h,\theta}$ and $Q_{h,\theta}$ is defined in \eqref{maxent} \;
    Learn the baseline function $t_{\phi}(\cdot)$ by optimizing $\phi$
    \begin{eqnarray}
        J_2(\phi)&:=&\sum_{(\bm{s}, \bm{a}, \bm{s}')\in\mathcal{D}\cup\mathcal{D}_R}l(Q_{h,\theta}(\bm{s},\bm{a})+t_{\phi}(\bm{s}) \nonumber \\
        &&-\gamma\max_{\bm{a}'\in\mathcal{A}}(Q_{h,\theta}(\bm{s}',\bm{a}')+t_{\phi}(\bm{s}'))) \label{j2}
    \end{eqnarray} \
    Then pass the auxiliary reward function $r_a$ \eqref{aux_rew} to the RL agent \;
    RL agent employs any RL framework using the modified reward function $r_e(s,a)+\beta(e)r_a(s,a)$.

    \label{algo:proposed}
\end{algorithm}

\vspace*{-10pt}
\section{Evaluation}
\label{sec:results}

\begin{figure*}
\centering
\vspace*{-25pt}
\subfigure[Decoding ErrPs]{
\begin{minipage}[t]{0.3\linewidth}
\centering
\includegraphics[width=0.95\columnwidth]{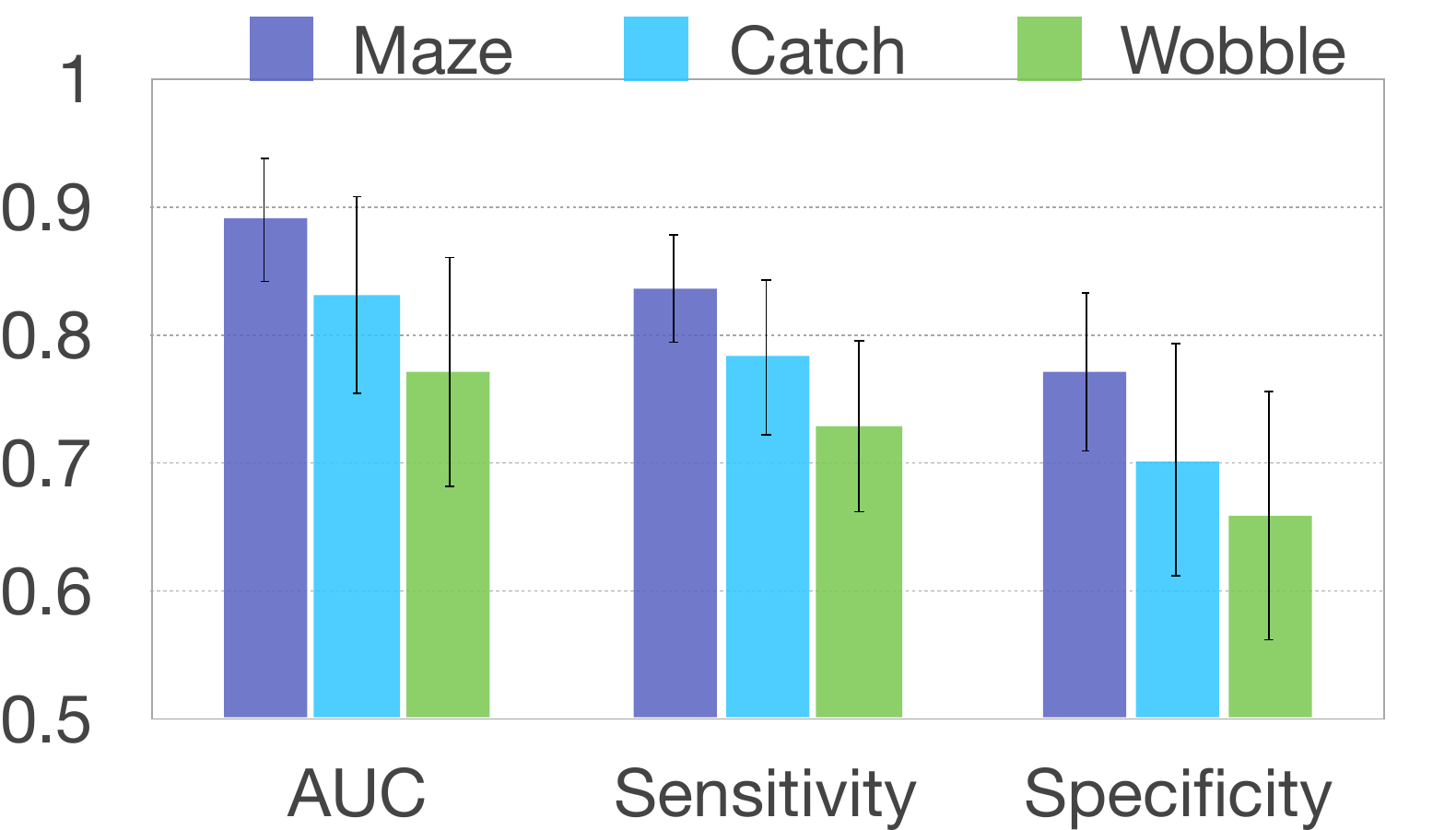}
\vspace*{-10pt}
\label{fig:perf_kfold}
\end{minipage}
}
\subfigure[Learning Curve]{
\begin{minipage}[t]{0.25\linewidth}
\centering
\includegraphics[width=0.95\columnwidth]{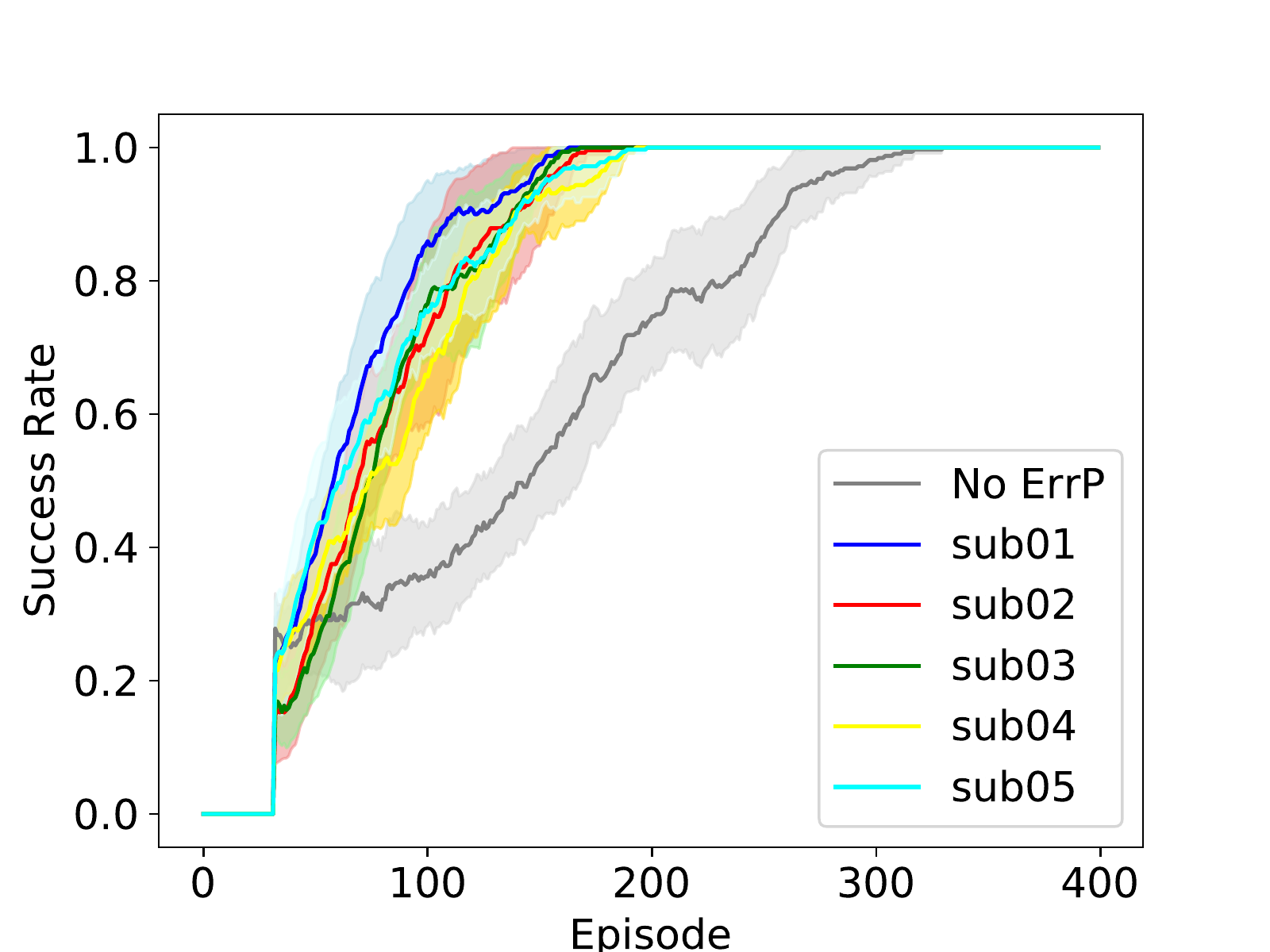}
\end{minipage}
}
\subfigure[Complete Episode]{
\begin{minipage}[t]{0.25\linewidth}
\centering
\includegraphics[width=0.95\columnwidth]{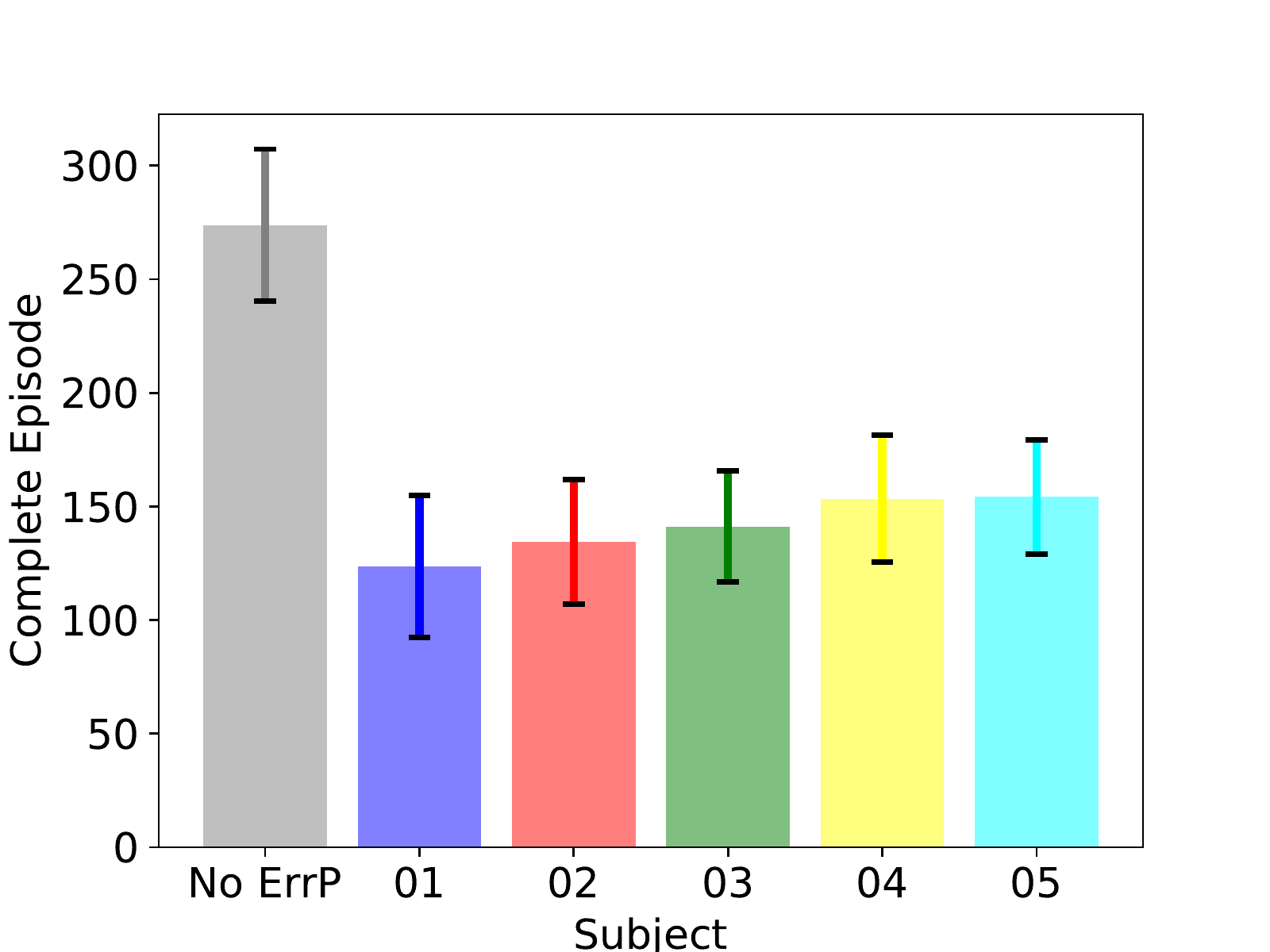}
\end{minipage}
}
\vspace*{-10pt}
\caption{Full Access method results: (a) 10-fold CV performance of each game without any zero-shot learning, (b) and (c) RL with \textit{full access} to ErrP feedback}
\vspace*{-10pt}
\label{fig:rl_errp}
\end{figure*}

\subsection{ Naive Approach}
\label{sec:resultsNaive}
We first validate the feasibility of decoding error-potentials using a 10-fold cross-validation scheme for each game. In this scheme, we split the state-action pairs of a game in 10-folds for training and testing of the ErrP decoder. In Figure \ref{fig:perf_kfold}, we show the performance of three games in terms of Area Under Curve (AUC) score, sensitivity and specificity, averaged over 5 subjects. The Maze game has the highest AUC score (0.89 $\pm$ 0.05) followed by Catch (0.83 $\pm$ 0.08) and Wobble (0.77 $\pm$ 0.09). 

\noindent 
\label{sec:full_access_perf}
As discussed in section \ref{subsec:full_access}, the \textit{full access} method is the most preliminary approach to incorporate implicit human feedback (in the form of decoded error-potentials) into the DRL model. 
It asks the external oracle (human) for the implicit feedback in every training step, reaching the maximum number of possible queries. Hence it has the fastest training convergence rate. We use this method as a benchmark for comparing the sample efficiency of the proposed RL framework. 
The evaluation metric adopted here is {\it success rate}, which is the ratio of success plays in the last 32 episodes. The training converges and terminates at {\it complete episode}, when the success rate reaches to 1. The results with real ErrP data of 5 subjects are shown in Figure \ref{fig:rl_errp}(b,c). 
We can see there is a significant improvement in the training convergence with all subjects when ErrP used.  
Here, \textit{No ErrP} refers to the BDQN performance without integrating the human feedback. In all plots of this paper, solid lines are average values over 10 random seeds, and shaded regions correspond to one standard deviation. We use BDQN (as introduced in section \ref{subsec:bdqn}) as the DRL model for all experiments conducted in this paper. However, the ErrP feedback here can be used to augment any RL algorithm. 

\subsection{Evaluation of the Proposed Solution}
In this subsection, we evaluate the performance of proposed approaches to practically integrate the implicit human feedback (via EEG) into the DRL algorithms. In addition, we provide subjective analysis of ErrP decoding errors and ablation study of the proposed reward shaping method.

\subsubsection{Zero-shot learning of ErrPs}
\label{sec:results0shot}

\begin{figure}
\centering
\subfigure[Over subjects]{
\begin{minipage}[t]{0.65\columnwidth}
\centering
\includegraphics[width=0.99\columnwidth]{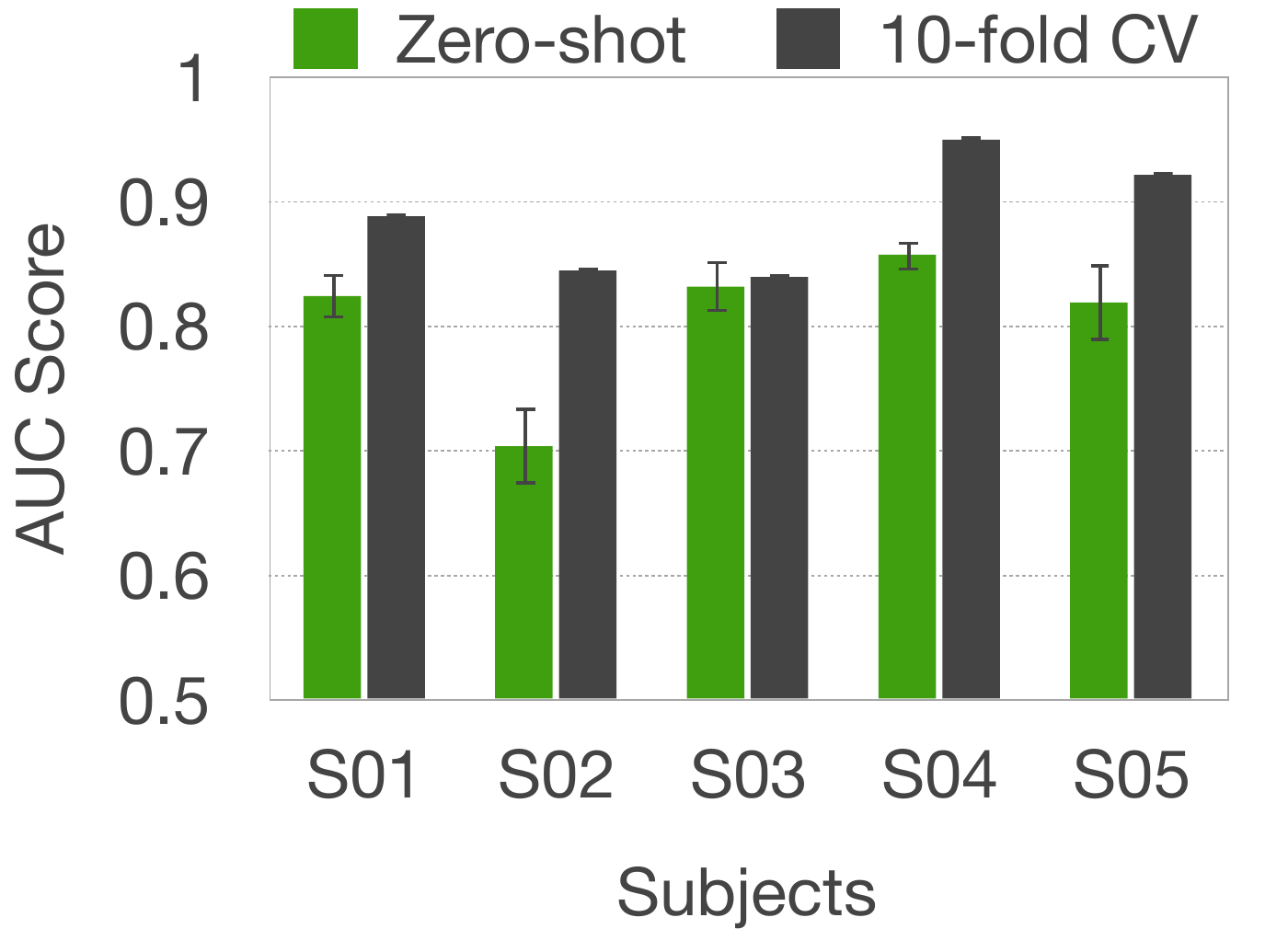}
\vspace*{-10pt}
\label{fig:perf_gmaze}
\end{minipage}
}

\subfigure[Over games]{
\begin{minipage}[t]{0.65\columnwidth}
\centering
\includegraphics[width=0.99\columnwidth]{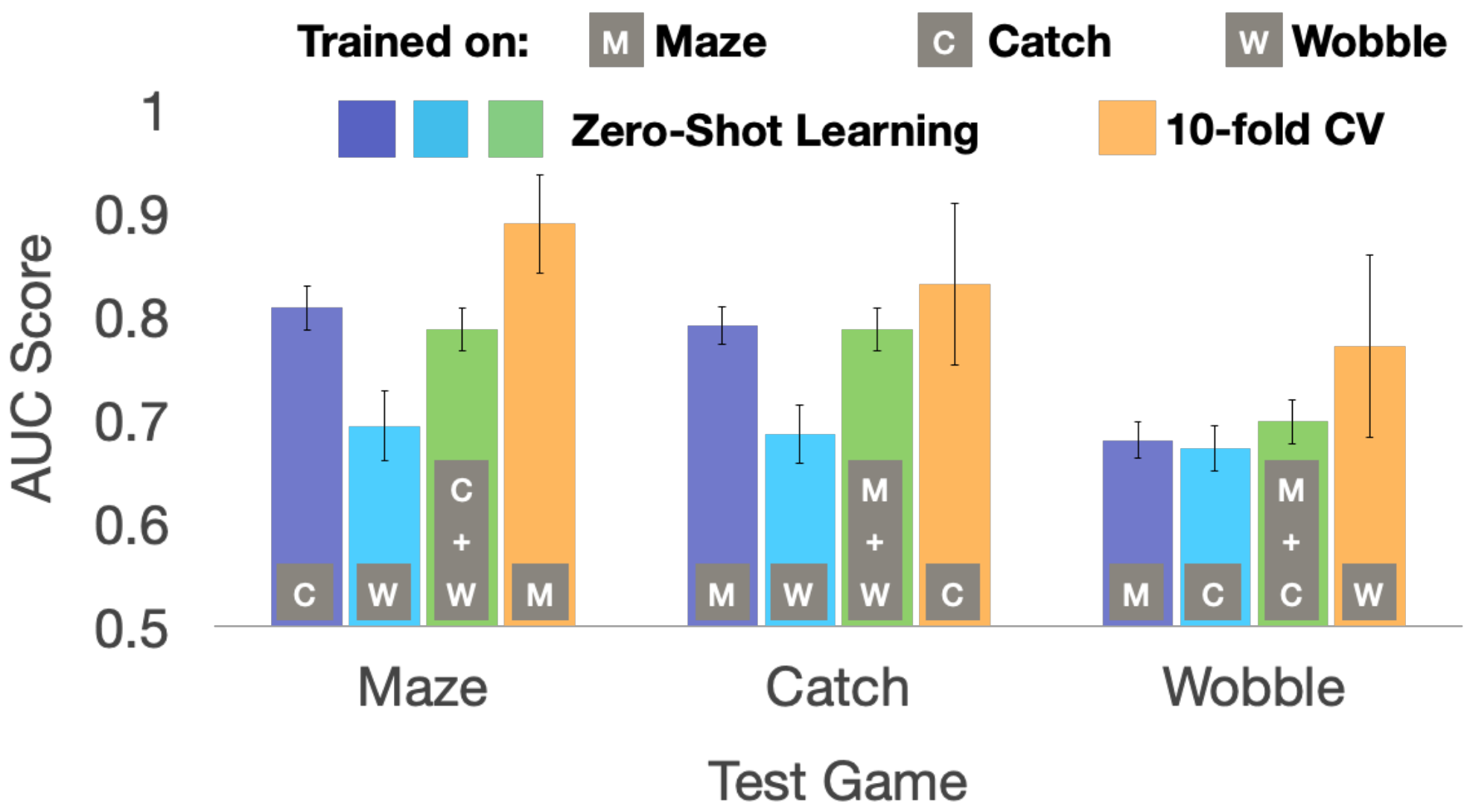}
\vspace*{-10pt}
\label{fig:perf_general}
\end{minipage}
}
\vspace*{-10pt}
\caption{Zero-shot learning of ErrP: (a) from Catch to Maze over subjects compared with 10-fold CV, (b) over all combinations of three games compared with 10-fold CV.}
\vspace*{-10pt}
\label{fig:errp_perf}
\end{figure}

\begin{figure*}[t]
\vspace*{-15pt}
\centering
\begin{minipage}[t]{0.27\linewidth}
\centering
\includegraphics[width=1.0\columnwidth]{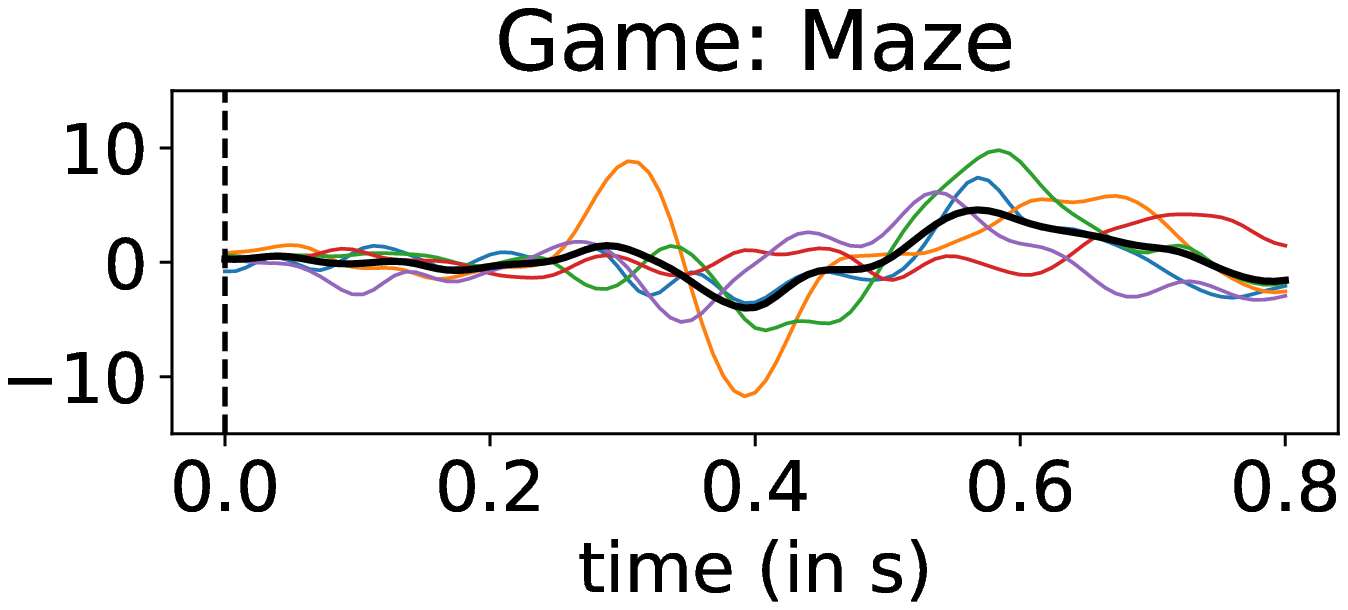}
\label{fig:waveform_maze}
\end{minipage}
\begin{minipage}[t]{0.27\linewidth}
\centering
\includegraphics[width=1.0\columnwidth]{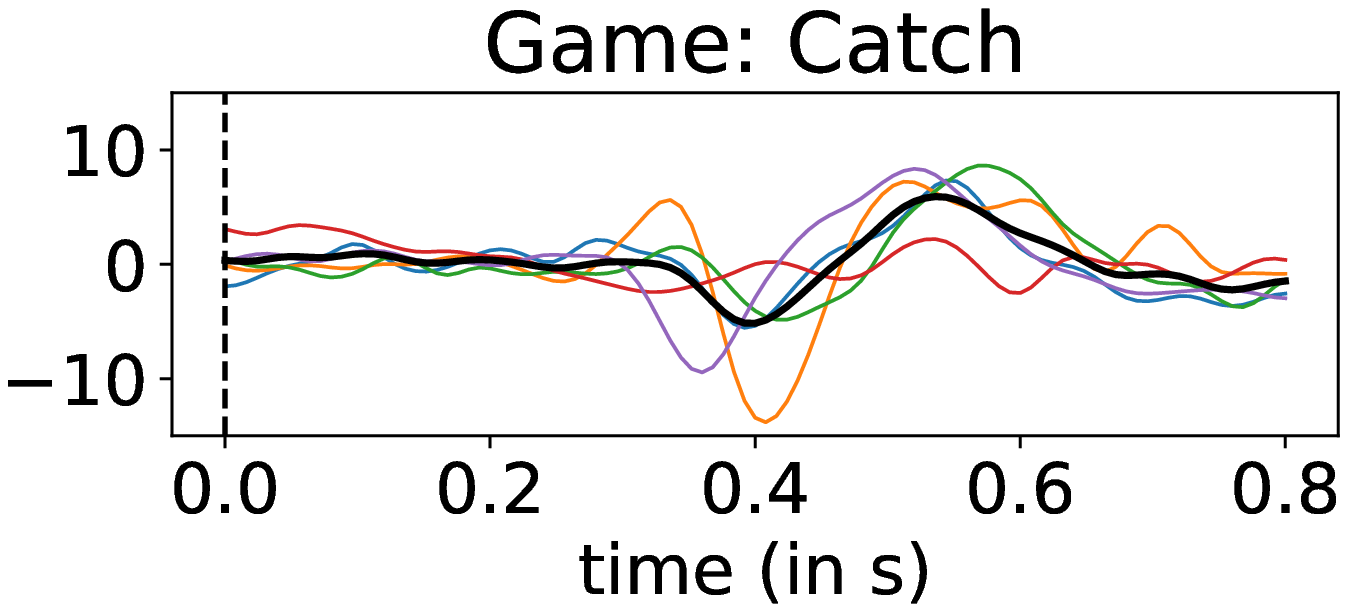}
\label{fig:waveform_catch}
\end{minipage}
\begin{minipage}[t]{0.27\linewidth}
\centering
\includegraphics[width=1.0\columnwidth]{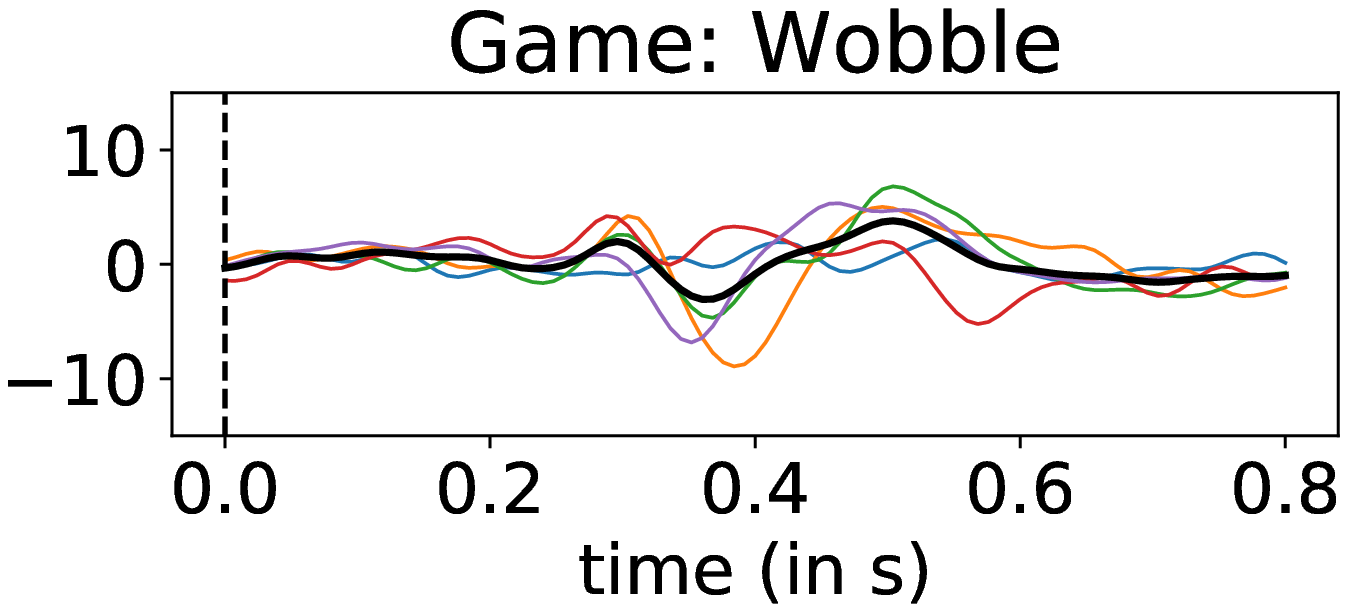}
\label{fig:waveform_wobble}
\end{minipage}
\begin{minipage}[t]{0.09\linewidth}
\centering
\vspace*{-75pt}
\includegraphics[width=1.0\columnwidth]{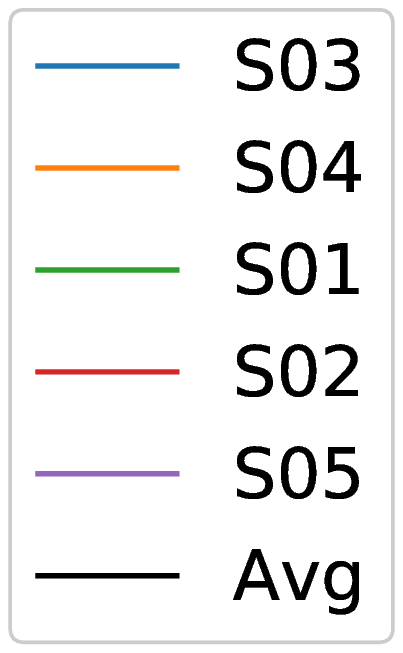}
\end{minipage}
\vspace*{-10pt}
\caption{Manifestation of error-potentials in time-domain: Grand average potentials (error-minus-correct conditions) are shown for Maze, Catch and Wobble game environments. Thick black line denotes the average over all the subjects. 
}
\vspace*{-5pt}
\label{fig:errp_waves}
\end{figure*}


Error-potentials in the EEG signals are studied under two major paradigms in human-machine interaction tasks, (i) \textit{feedback and response ErrPs:} error made by human \cite{carter1998anterior,falkenstein2000erp,blankertz2003boosting,parra2003response}, (ii) \textit{interaction ErrPs:} error made by machine in interpreting human intent \cite{ferrez2005you}. Another interesting paradigm is \textit{observation ErrPs}, when a human is watching (and silently assessing) the machine performing a specific task \cite{chavarriaga2010learning}. {We make the case for the generalizability of these ErrPs owing to their universality across humans and other primates in section \ref{sec:errp_case}\ref{subsec:errp_evo}. We also observe that} the manifestation of these potentials across these paradigms are found quite similar in terms of their general shape, negative and positive peak latency, and frequency characteristics\cite{ferrez2005you,chavarriaga2010learning}. This prompts us to explore the consistency of the error-potentials across different environments (i.e., games, in our case) within the \textit{observation ErrPs}. 

In Figure \ref{fig:errp_waves}, we plot the grand average waveforms across three environments (Maze, Catch and Wobble), to visually validate the consistency of potentials. We can see that the shape of negativity, and the peak latency is quite consistent across the three game environments. {These indicators prompt us to experimentally explore the domain of zero-shot learning for ErrPs in evaluation section \ref{sec:results0shot}. We show that by training error potentials on one game, we are able to cover the variability of error potentials in other games as well which suggests that error-potentials are indeed generalizable across environments, and can further be used to inform deep reinforcement learning algorithm in new and unseen environments.} 

To evaluate the zero-shot learning capability of error-potentials and the decoding algorithm, we train on the samples collected from the Catch game and test on the Maze game. \textcolor{new}{As Catch is a simple game, we assume the optimal action for each state is already known (providing the labeled examples to train the ErrP decoder)}. 
However, the Maze game needed to be solved, hence, we do not make any assumptions about the optimality of the actions. 
\textcolor{new}{In Figure \ref{fig:perf_gmaze}, we provide the zero-shot learning performance and compare it against  the 10-fold Cross-Validation (CV) scheme discussed in section \ref{sec:resultsNaive}}. Further, we present the AUC score of \textcolor{new}{zero-shot learning performance} over all training and testing combinations in Figure \ref{fig:perf_general}. \textcolor{new}{We use the Area Under Curve (AUC) as the performance metric for the decoding of error-potentials.} We can see that the ErrPs recorded for Catch game, are able to capture more than 80\% of the variability in the ErrPs for Maze game. \textcolor{new}{Averaged over 5 subjects, the decoder performs with an AUC score of 0.8078 ($\pm 0.022$) when trained on the Catch game. This compared with the performance of 0.693 ($\pm 0.034$) when trained using Wobble labels. Similarly, Catch and Wobble performs with an average AUC score of 0.790 ($\pm 0.018$) and 0.680 ($\pm 0.018$) respectively, when trained on labels obtained through the Maze environment. These experiments validate that the error-potentials can be learned in a zero-shot manner to avoid re-training of the human feedback (via EEG) decoder.}

\subsubsection{Evaluation of Robust Reward Shaping with Human ErrP}
\label{sec:second_evaluation}
\textcolor{new}{For the evaluation of Algorithm 2, we generated stochastic trajectories for the Maze game}
 by Monte Carlo Tree Search (MCTS) discussed above, where the trade-off parameter $c$ is set to $0.5$. Before training the RL agent, each \textcolor{new}{human} subject \textcolor{new}{provided implicit feedback (via ErrP) as explained in the experimental protocol (section \ref{subsec:experiment})} on every state-action pair along these trajectories. \textcolor{new}{We evaluated the performance of the proposed approach with 10 and 20 initial trajectories, in the demonstrations in Figure \ref{fig:imperfect}, each for 5 subjects. We use the Bayesian DQN as the DRL model.}
\begin{figure}
\centering
\subfigure[10 Trajectories]{
\begin{minipage}[t]{0.65\columnwidth}
\centering
\includegraphics[width=1.08\columnwidth]{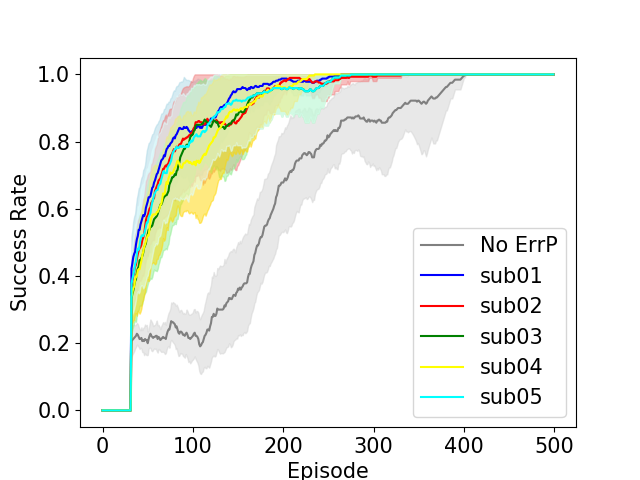}
\end{minipage}
}

\subfigure[20 Trajectories]{
\begin{minipage}[t]{0.65\columnwidth}
\centering
\includegraphics[width=1.08\columnwidth]{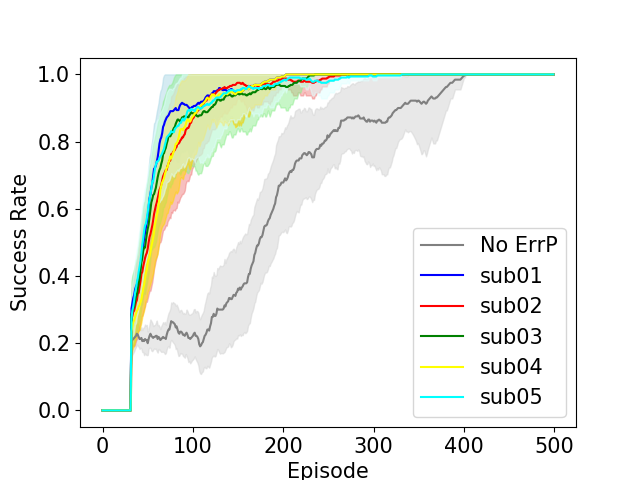}
\end{minipage}
}
\vspace*{-10pt}
\caption{Evaluation of the proposed reward shaping method.}
\label{fig:irl_errp}
\end{figure}

\begin{table}
\caption{Average Number of Queries on Maze Game}
\setlength{\tabcolsep}{1.5mm}{\small
\begin{tabular}{c|c|c|c|c|c}
\toprule
Subject & 01 & 02 & 03 & 04 & 05 \\
\midrule
Full access & $1879.4$ & $2072.1$ & $2293.7$ & $1975.4$ & $2130.1$ \\
Proposed method &  $505.7$ & $394.7$ & $587.1$ & $681.4$ & $361.3$ \\
\bottomrule
\end{tabular}}
\label{tab:maze}
\end{table}

The acceleration of RL due to \textcolor{new}{human feedback} is shown in Figure \ref{fig:irl_errp}(a) \textcolor{new}{for 10 trajectories}, where the base model is Bayesian DQN. We can see the significant acceleration in training convergence \textcolor{new}{ in Figure \ref{fig:irl_errp}(a) in terms of the \textit{success rate} for 5 subjects and compared against the case of \textit{No ErrP, i.e. no human feedback}. Subject 01 has the highest fidelity for error-potentials, and hence, RL algorithm converges at much faster rate when relies upon the feedback obtained by Subject 01.} \textcolor{new}{It is evident from the results that the error-potential decoding performance is sufficient to achieve around 2x improvement in training time (in terms of the number of episodes required).}
\textcolor{new}{Similarly, Figures \ref{fig:irl_errp}(b) shows the \textit{success rate} and convergence curve for training to complete, for 20 trajectories.}
Comparing Figures \ref{fig:irl_errp}(a) and (b), we can see that the \textcolor{new}{training converges at much faster rate when the number of initial trajectories are increased. Further,} 
the learning variance also decreases with more trajectories. The comparison between Figure \ref{fig:irl_errp} and Figure \ref{fig:rl_errp}(c) shows that the proposed framework learns faster than No-ErrP case, while outperforming the \textit{full access} case, even though \textit{full access} \textcolor{new}{requires significantly larger amount of queries}. \textcolor{new}{We also compare the number of ErrP queries for \textit{full access} and proposed method} in Table \ref{tab:maze}, according to the statistics on experiments with 20 trajectories. \textcolor{new}{On an average for 5 subjects, the proposed approach makes 75.56\% less queries as compared to the \textit{full access}}. As \textit{full access} queries for feedback label at every learning step, i.e. state-action pair, while the proposed framework queries only on the trajectories given initially, the total number of queries made are significantly reduced.

\begin{table}
\centering
\vspace{-5pt}
\caption{Accuracy and standard deviations per subject for ErrP and non-ErrP SAPs}
\setlength{\tabcolsep}{1.5mm}{\footnotesize
\begin{tabular}{c|c|c|c|c}
\toprule
Subject & ErrP SAPs  & ErrP SAPs & non-ErrP   & non-ErrP  \\
 & mean  & std dev & SAPs mean  &  SAPs std dev \\
\midrule

S12 &0.79 &0.27 &0.75 &0.17 \\

S07 &0.8 &0.3 &0.85 &0.16 \\

S02 &0.73 &0.29 &0.77 &0.15 \\

S08 &0.6 &0.25 &0.56 &0.14 \\

S01 &0.8 &0.25 &0.77 &0.16 \\

S04 &0.78 &0.25 &0.63 &0.16 \\

S16 &0.73 &0.3 &0.78 &0.14 \\

S03 &0.65 &0.25 &0.61 &0.13 \\

S06 &0.73 &0.3 &0.64 &0.17 \\

S05 &0.75 &0.3 &0.72 &0.13 \\

S09 &0.71 &0.27 &0.66 &0.13 \\

S15 &0.67 &0.31 &0.65 &0.1 \\
\midrule
Average &0.73 & 0.28 &0.70 & 0.14\\
\bottomrule
\end{tabular}}
\label{tab:maze_per_sub_acc}
\end{table}

\color{mohit}
\subsubsection{Analysis of the dependence and subjectivity of errors} 
\begin{figure}
\centering
\subfigure[ErrP SAPs accuracy mean]{
\begin{minipage}[t]{0.44\columnwidth}
\centering
\includegraphics[width=1.08\columnwidth]{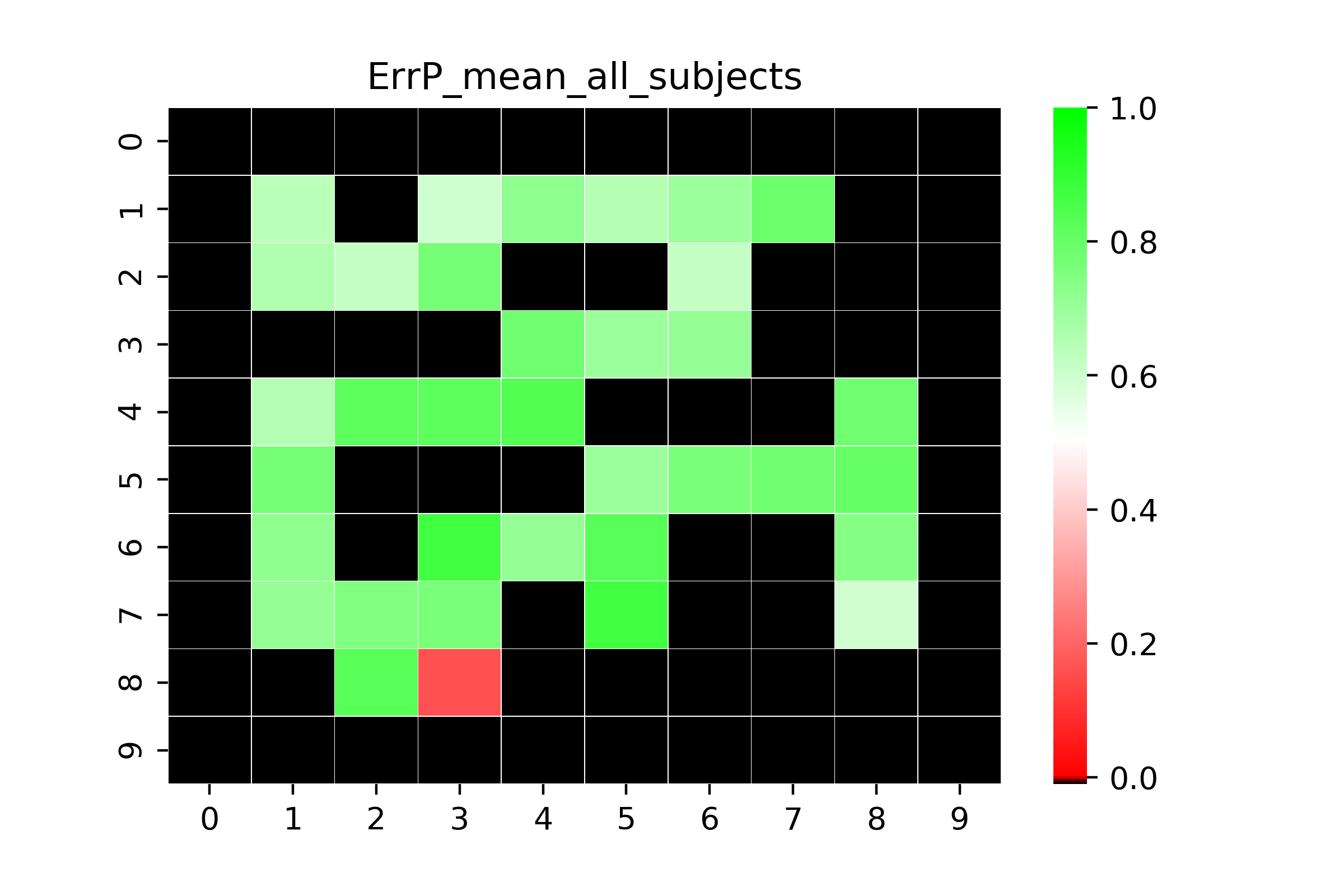}
\end{minipage}
}
\subfigure[ErrP SAPs accuracy std deviation]{
\begin{minipage}[t]{0.44\columnwidth}
\centering
\includegraphics[width=1.08\columnwidth]{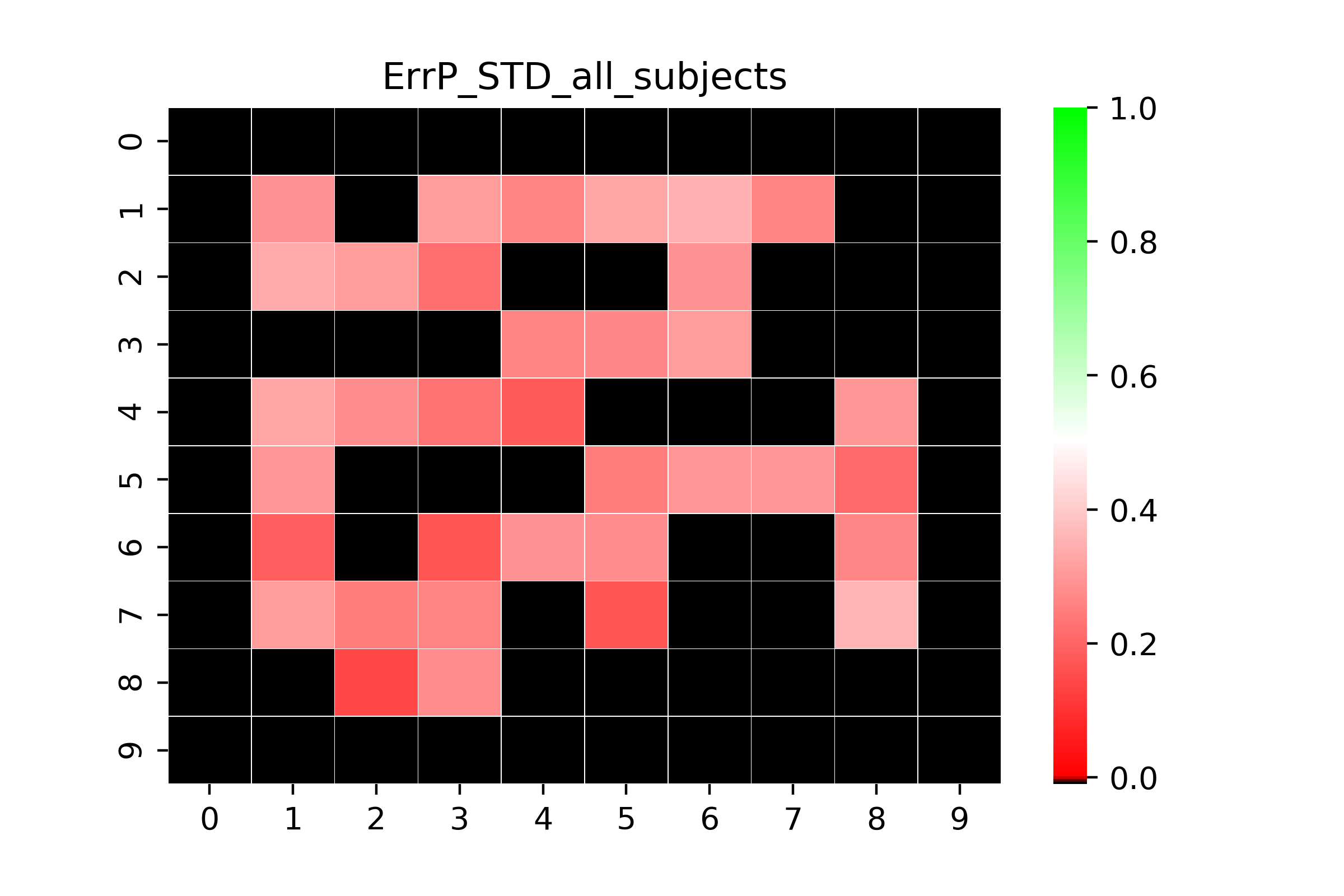}
\end{minipage}
}
\subfigure[non-ErrP SAPs accuracy mean]{
\begin{minipage}[t]{0.44\columnwidth}
\centering
\includegraphics[width=1.08\columnwidth]{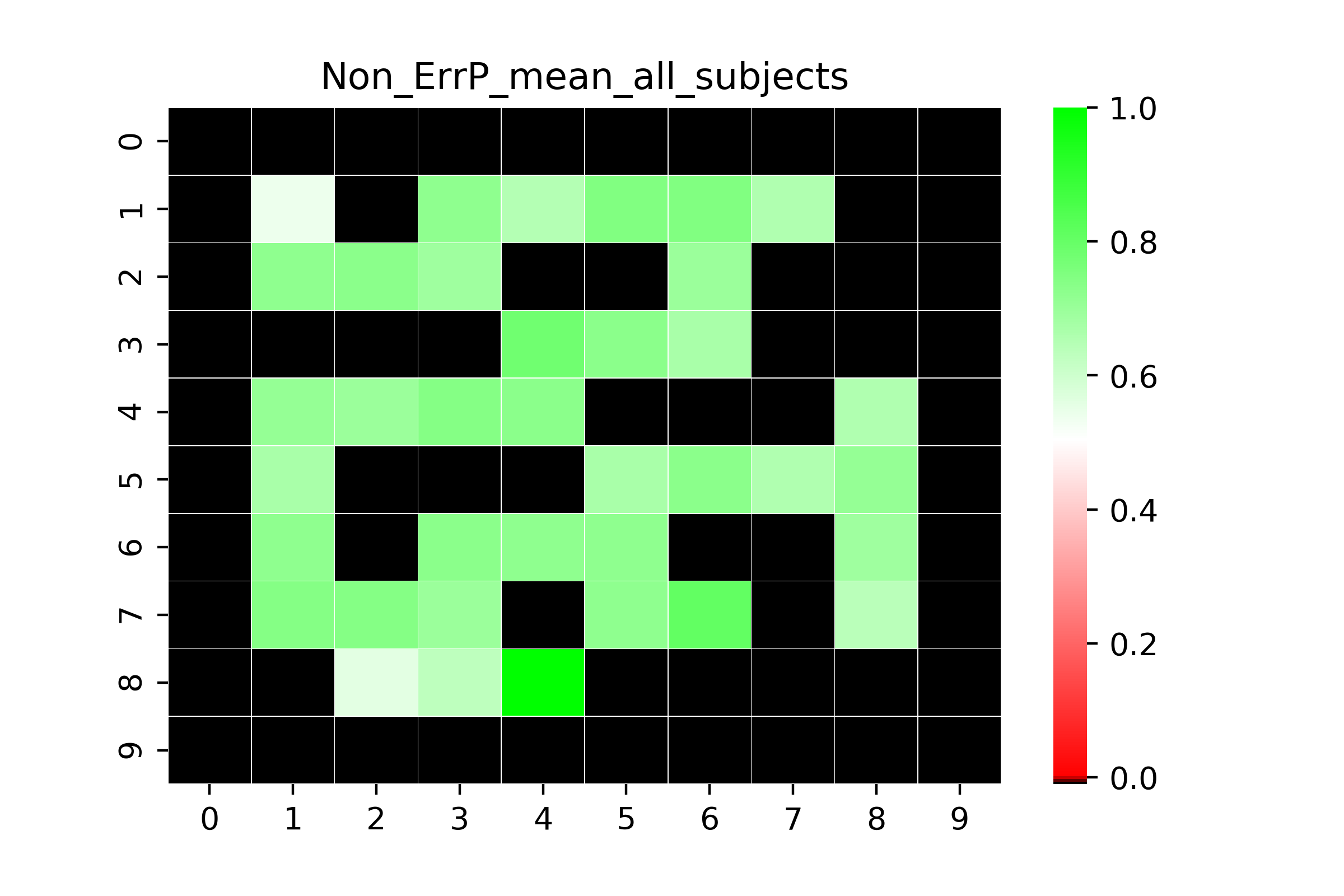}
\end{minipage}
}
\subfigure[non-ErrP SAPs accuracy std deviation]{
\begin{minipage}[t]{0.44\columnwidth}
\centering
\includegraphics[width=1.08\columnwidth]{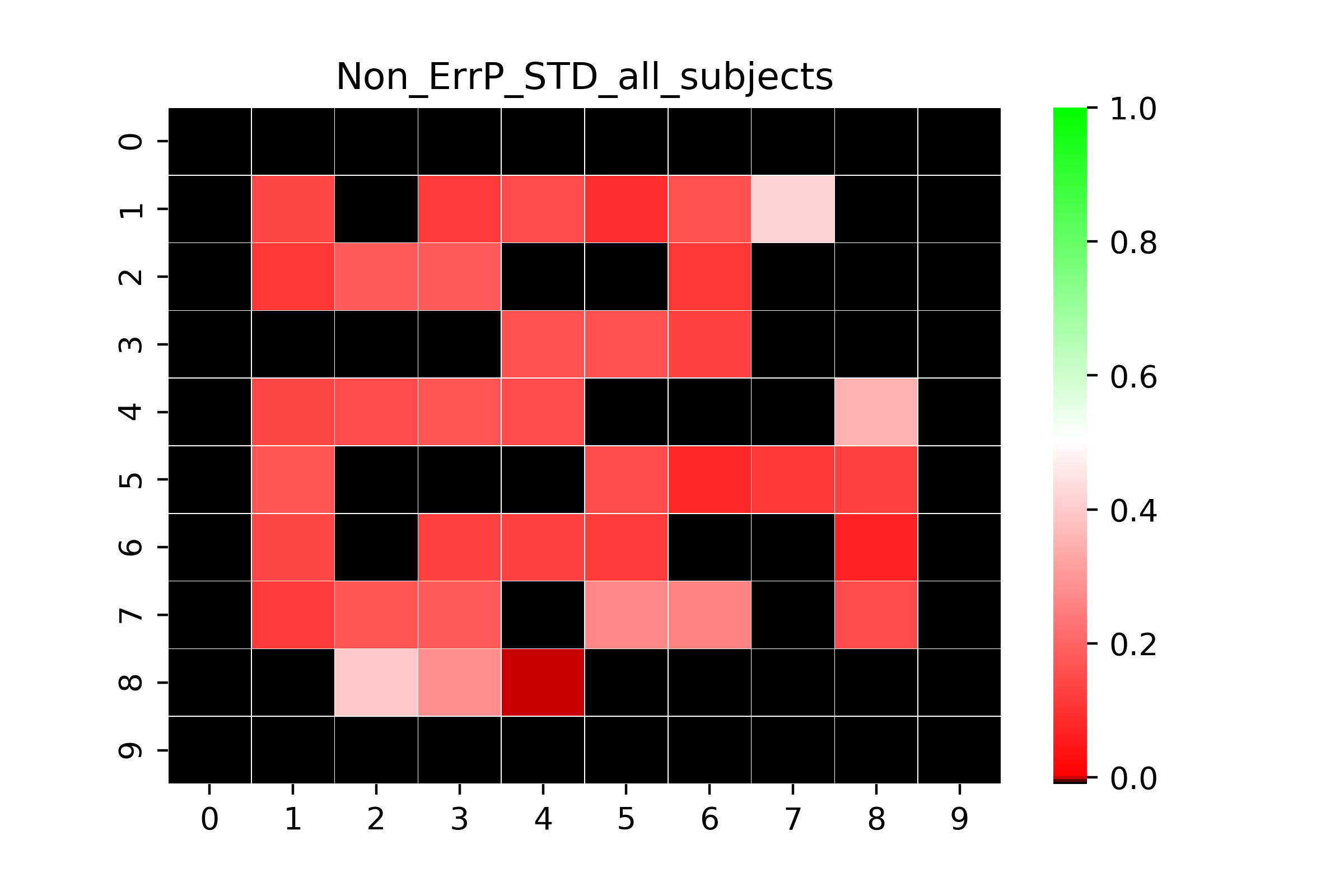}
\end{minipage}
}
\caption{Differences between ErrP and non-ErrP accuracies for each initial state over all users}
\vspace*{-10pt}
\label{fig:errp_non_errp_mean_std}
\end{figure}

In this section, we analyze the detection accuracy of error-potentials for the Maze game, to develop insights into the characteristics of error-potential based on the users and provided stimulations. The EEG samples recorded for the Maze experiment can be presented along two independent dimensions, (i) users and (ii) state-action pair of the agent (i.e., stimulation). Within the state-action pairs (or SAPs for short), if the action is correct, it is called a \textit{non-ErrP SAP}, otherwise an \textit{ErrP SAP}. Please note that the term \textit{non-ErrP state-action pair} or \textit{non-ErrP SAP} refers to a correct action taken by the agent given a state (due to the expected absence of an ErrP response in the brain), and does not refer to a system where we do not use implicit human feedback using EEG. 
\begin{itemize}
    \item \textbf{Experiment 1: Subjectivity over correct and incorrect actions.} For each user, we divide the EEG trials into two categories (a) \textit{ErrP SAPs}, and (b) \textit{non-ErrP SAPs}. For each user and category, we compute the mean and standard deviation of classification accuracy of EEG trials, and present in Table \ref{tab:maze_per_sub_acc}. We can observe that the per user standard deviations for ErrP SAPs is roughly double the standard deviations for non-ErrP SAPs. The aggregate per user standard deviation across the ErrP SAPs is 0.28 and 0.14 for non-ErrP SAPs. This difference in per user standard deviations is statistically significant (p$<$0.001). We also calculate the standard deviations across our user accuracy vectors for both ErrP SAPs and non-ErrP  SAPs and find that the standard deviations for the per user accuracy vectors are 0.06 and 0.08 respectively.
    \item \textbf{Experiment 2: Subjectivity over users.}  In this experiment, for each unique state-action pair, we average the performance of EEG trials of all users. We achieved a mean and standard deviation of 0.75 and 0.13, and 0.75 and 0.07 for \textit{ErrP SAPs} and \textit{non-ErrP SAPs} respectively. We use Levene's test \cite{levene1960robust} to conclude that the difference in variance between these two population samples is statistically significant ($p=0.023 < 0.05$).
    \item \textbf{Experiment 3: Subjectivity over states.} For each unique state in Maze game, we plot the mean and standard deviation of EEG trial performance in Fig. \ref{fig:errp_non_errp_mean_std}. We plot the classfier accuracy for ErrP SAPs and non-ErrP SAPs respectively based on their initial state on the maze. We can visualize that the plot corresponding the standard deviation for non-ErrP SAPs is darker (indicating lower standard deviation) compared to the plot corresponding to the deviations for ErrP SAPs. We can also see that within a plot, there is also a gradation in the accuracy (indicated by different shades of green) implying that there is some dissimilarity among erroneous states and hence subjectivity on the user's part and diminishing the argument that erroneous vs non-erroneous scenarios are purely binary.
    \item \textbf{Experiment 4: Errors of commission and omission.} In this experiment, we consider only the erroneous actions (ErrP SAPs) and split the EEG trials into two categories, (i) commission errors and (ii) omission errors. We do this in order to better understand the impact on ErrPs, based on the type of error committed. A commission error is defined as an agent making an incorrect move to a new cell, while omission error refers to the incorrect action of agent by staying in the same cell grid. The total state-actions pairs for commission and omission are distributed fairly (out of 71 unique state-action pairs, 34 correspond to errors of omission and the remaining 37 correspond to errors of commission). However we observe that among the state-action pairs which had very high accuracies, state-action pairs corresponding to errors of commission are disproportionately represented. Out of the top 5 state-action pairs that have the highest accuracy, all of them represent errors of commission and out of the top 10 state-action pairs that have the highest accuracy, 9 of them signify errors of commission. This was also indicated by the fact that errors of omission had a mean accuracy of 72\% whereas errors of commission had a much higher mean accuracy of 77\%. This implies that the error scenarios that are the easiest to detect are likely to be errors of commission. This has certain implications that bolster the hypothesis that certain errors are indeed more ''valuable'' to a user than others and hence generate a far more noticeable response in the brain.
\end{itemize}
These 4 experiments collectively lead us to 2 main insights.

\textcolor{mohit}{
\begin{enumerate}[label=(\alph*)]
\item
Per subject, owing to the differences in variances, there is less variation in the non-ErrP accuracies compared to the ErrP accuracies implying that erroneous scenarios lead to more variation in the classifier accuracy and by extension, in the brain's response, than non-erroneous scenarios. This further implies that there is a gradation in error detection unlike it being a binary phenomenon which makes certain errors easier to detect and certain others more difficult to detect.
\item
The differences in variations in classifier accuracy between ErrP and non-ErrP SAPs diminishes when we average the accuracies over the SAPs and represent them as a function of users. This implies that the variation in the accuracy of ErrP vs non-ErrP is impacted more by differences in SAPs compared to the differences in users.
\end{enumerate}
}
\color{black}

\subsubsection{Robustness Evaluation}
\begin{figure}
\centering
\subfigure[Subject 02-10 Trajectories]{
\begin{minipage}[t]{0.70\columnwidth}
\centering
\includegraphics[width=1\columnwidth]{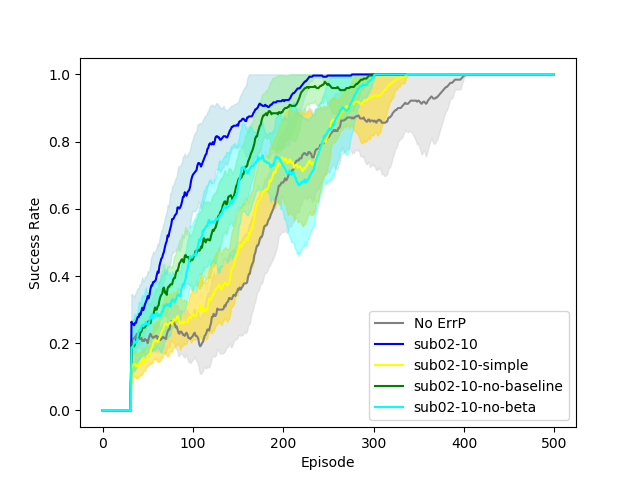}
\end{minipage}
}

\subfigure[Subject 02-20 Trajectories]{
\begin{minipage}[t]{0.70\columnwidth}
\centering
\includegraphics[width=1\columnwidth]{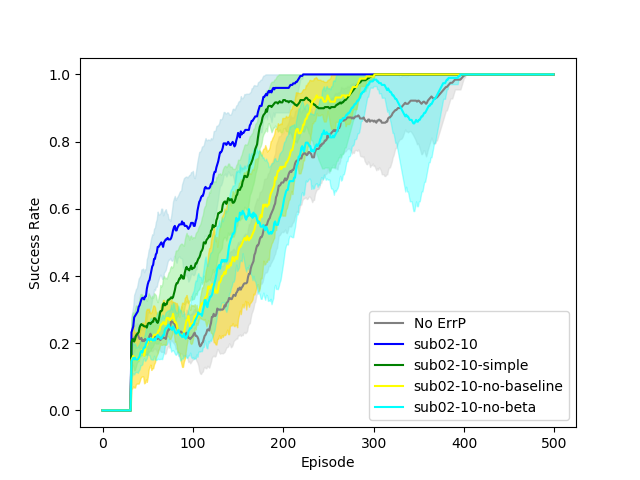}
\end{minipage}
}

\subfigure[Subject 07-10 Trajectories]{
\begin{minipage}[t]{0.70\columnwidth}
\centering
\includegraphics[width=1\columnwidth]{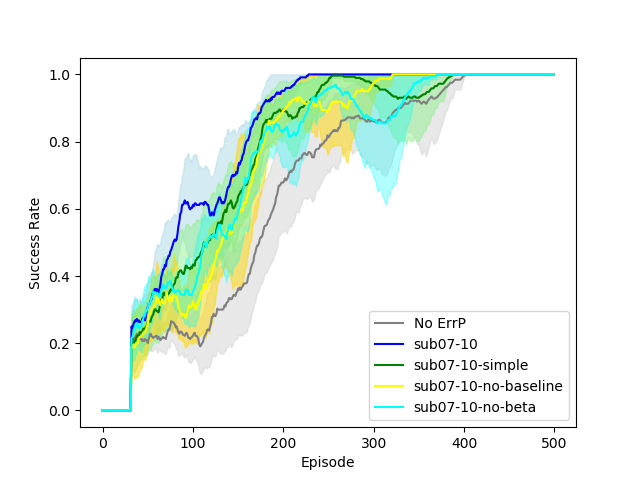}
\end{minipage}
}

\subfigure[Subject 07-20 Trajectories]{
\begin{minipage}[t]{0.70\columnwidth}
\centering
\includegraphics[width=1\columnwidth]{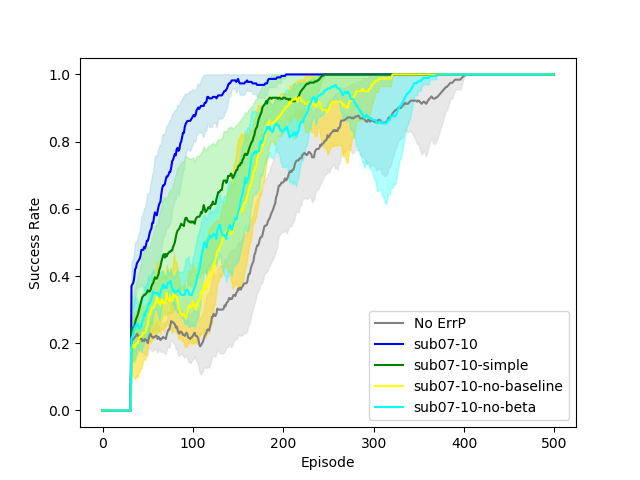}
\end{minipage}
}

\caption{Robustness and Ablation Study of the proposed reward shaping method. Simple: modeling human feedback simply by a neural network. No-baseline: method without function $t_{\phi}(\cdot)$. No-beta: method without combining coefficient $\beta(\cdot)$}
\label{fig:irl_robust}
\end{figure}

Because the generation process and decoding of brain signal are stochastic, the robustness to wrong ErrP labels is important when incorporating human feedback (via EEG) into reward shaping method. We are going to show that modeling the human policy as soft Q policy, as we did in (\ref{maxent}), can make the learned auxiliary function $r_a$ resist to wrong human feedback. In the comparison on robustness, the baseline method, called "{\it simple}", is to simply use a bootstrap neural network to generalize the binary ErrP labels across the state space, same as \cite{xiao2020fresh}. Both {\it simple} benchmark and the proposed robust reward shaping are trained on the same set of trajectories and human labels. The neural network in both methods is MLP, having two hidden layers of 64 units. And the number of bootstrap head in "simple" benchmark is set to 5. We evaluate both {\it simple} and the proposed methods on subject 02 and subject 07, whose accuracy are $0.71$ and $0.78$ respectively. The comparison result is shown in Figure \ref{fig:irl_robust}. We can see that the proposed method performs better in both subjects with different initial trajectories. That is because the proposed method treats the human feedback in a probabilistic way, and the baseline function $t$ can incorporate the state transition information to attenuate the influence of wrong human feedback. Moreover, the comparison of all cases shows that the performance gain of {\it simple} benchmark over no-ErrP method is decreased when the error probability of human label increases. 

\subsubsection{Ablation Study}
In this section, we conduct ablation study on the proposed robust reward shaping with human feedback. We first specifically evaluate the effect of baseline function $t$ learned from \eqref{j2}. Because the human feedback labels in the initial trajectories cannot cover the whole state space and some labels are wrong, the learned Q function of human $Q_h(\cdot,\cdot)$ may not be compatible with the state dynamics of the environment. Thus we introduce a baseline function only in terms of state to smoothen the learned Q function. Here the ablation evaluation on baseline function is still on subject 02 and 07, same as the section above. We realize $Q_h(\cdot, \cdot)$ and $t(\cdot)$ functions both by a two-layer MLP with 64 hidden units on each hidden layer and ReLU activation. We empirically find that it is enough to use 20 trajectories to train function $t(\cdot)$. The baseline is the proposed reward shaping method without $t$ in \eqref{aux_rew}, corresponding to {\it no-baseline} curves in Figure \ref{fig:irl_robust}, where the auxiliary reward function is only the Bellman difference of Q function between adjacent states. The comparison result is shown in Figure \ref{fig:irl_robust}. We can see that the baseline function can improve the convergence speed in all cases, and it can even do better than {\it simple} method in some cases, showing the importance of the baseline function here. 

In addition, we also conduct the ablation study on the combining coefficient $\beta(\cdot)$. The benchmark method is to directly sum auxiliary reward $r_a$ and environmental reward $r_e$ together. The coefficient in the proposed method is set to  $\beta(e)=3e^{-e/80}$. Comparison result is shown in Figure \ref{fig:irl_robust} where {\it no-beta} curves corresponding benchmarks on combining coefficients. We can see this exponentially decreasing coefficient can stabilize the training process significantly, and hence improve the convergence speed.

\section{Conclusions and Future Work}
\label{sec:discussion}

In this work, we investigated an interesting paradigm to obtain and integrate the implicit human feedback with RL algorithms. We first demonstrated the feasibility of obtaining implicit human feedback by capturing error-potentials of a human observer watching an agent learning to play several different visual-based games, and then decoding the signals appropriately and using them as an auxiliary reward function to help an RL agent. Then we argued that the definition of ErrPs could be learned in a zero-shot manner across different environments, eliminating the need of re-training over new and unseen environments. We validated the acceleration in learning of games through augmenting the RL agent by ErrP feedback using a naive approach, i.e., \textit{full access} method. We then proposed a novel RL
framework, improving the label efficiency and reducing human cognitive load. We experimentally showed that the proposed RL framework could accelerate the training of RL agent by 2.25x, while reducing the number of queries required by 75.56\%.



\bibliographystyle{plain}
\bibliography{ijcai20, iclr2020_conference, rl}

\end{document}